\documentclass[10pt,twocolumn,letterpaper]{article}

\usepackage{cvpr}              %

\usepackage{algorithm} 
\usepackage{algpseudocode}
\usepackage[whole]{bxcjkjatype}

\usepackage{booktabs}
\usepackage{comment}

\usepackage{booktabs}
\usepackage{caption}
\usepackage{makecell}
\usepackage{fancyvrb}

\usepackage{tabularx}

\usepackage{multirow}
\usepackage[export]{adjustbox}
\usepackage{array,makecell,booktabs}
\usepackage{fvextra}

\definecolor{cvprblue}{rgb}{0.21,0.49,0.74}
\usepackage[pagebackref,breaklinks,colorlinks,allcolors=cvprblue]{hyperref}

\def\paperID{} %
\def\confName{CVPR}
\def\confYear{2026}

\title{SLVMEval: Synthetic Meta Evaluation Benchmark \\for Text-to-Long Video Generation}

\author{
  Ryosuke Matsuda$^{1}$ \quad
  Keito Kudo$^{1}$ \quad
  Haruto Yoshida$^{1}$ \quad
  Nobuyuki Shimizu$^{2}$ \quad
  Jun Suzuki$^{1}$ \\
  $^{1}$Tohoku University \quad
  $^{2}$LY Corporation \quad \\
  \href{https://slvmeval.github.io/}{https://slvmeval.github.io/}
}

\begin{document}
\maketitle

\begin{abstract}
This paper proposes the synthetic long-video meta-evaluation (SLVMEval), a benchmark to perform meta-evaluations of text-to-video (T2V) evaluation systems. 
The proposed SLVMEval benchmark focuses on assessing these systems on videos of up to 10,486 s (approximately 3 h).
The benchmark targets a fundamental requirement, i.e., whether the systems can accurately assess video quality in settings that are easy for humans to assess.
We adopt a pairwise comparison-based meta-evaluation framework.
Building on dense video-captioning datasets, we synthetically degrade source videos to create controlled “high-quality versus low-quality” pairs across 10 distinct aspects. 
Then, we employ crowdsourcing to filter and retain only those pairs in which the degradation is clearly perceptible, thereby establishing an effective final testbed.
Using this testbed, we assess the reliability of existing evaluation systems in ranking these pairs. 
Experimental results demonstrate that human evaluators can identify the better long video with 84.7\%--96.8\% accuracy, and in nine of the 10 aspects, the accuracy of these systems falls short of the human assessment, which reveals weaknesses in text-to-long video evaluation.

\end{abstract}

\section{Introduction}
\label{sec:intro}
In contemporary society, much of the video content we consume daily, especially movies and TV series, is long-form content typically spanning tens of minutes to several hours.
Despite recent advances in text-to-video (T2V) models~\cite{GoogleDeepMind2025Veo3,openai2024videosim}, their outputs remain limited to a few minutes, and generating videos at movie scale length faces technical challenges.
In response, systems that can, in principle, generate arbitrarily long videos have recently emerged~\cite{villegas2023phenaki,chen2025skyreelsv2infinitelengthfilmgenerative}, highlighting long-form video generation as the next frontier.

Despite the momentum in text-to-long video (T2LV) research, validation of functional evaluation metrics for T2LV is lacking. 
Currently, researchers frequently employ the VideoScore~\cite{he-etal-2024-videoscore} metrics to evaluate T2LV results.
However, such metrics were originally designed for short videos lasting only a few seconds to tens of seconds; thus, the T2LV results may not be evaluated appropriately due to a mismatch in video length.
In addition, even if we attempt to validate whether such mismatch occurs using human-annotated datasets, e.g., VBench~\cite{huang2023vbench} and UVE~\cite{liu2025uvemllmsunifiedevaluators}, these datasets cannot reliably assess the evaluation metrics used for T2LV because they focus exclusively on short videos.
From this perspective, developing a meta-evaluation benchmark for T2LV is a crucial step toward advancing T2LV research.

\begin{figure}[t]
  \centering
  \adjincludegraphics[width=\linewidth,clip,trim={0 {.29\height} 0 {0}}]{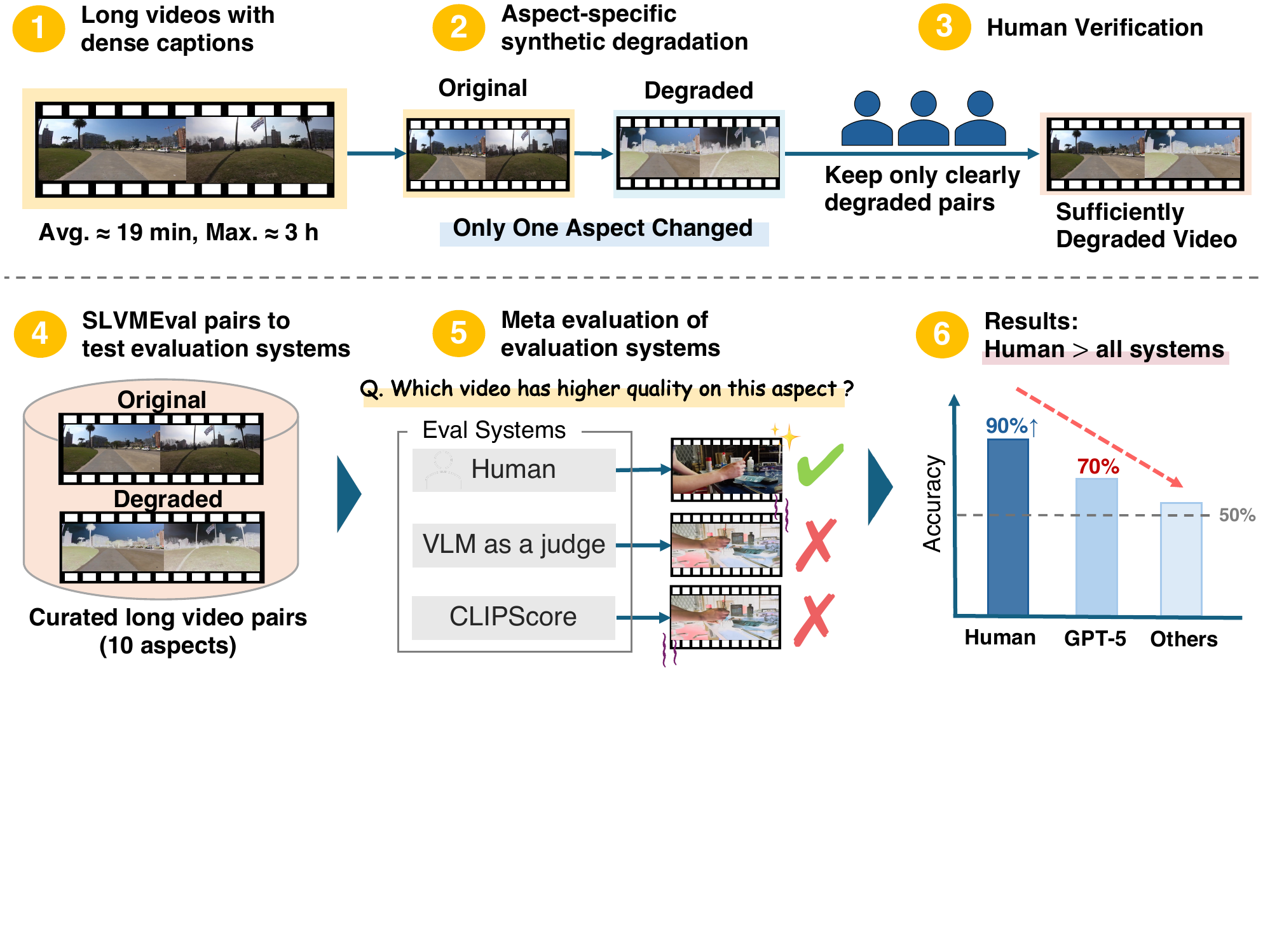}
  \caption{
    Overview of proposed SLVMEval benchmark. We construct human-validated pairs of original and aspect (specifically degraded long videos), and we test various automatic evaluation systems. Human evaluators reliably pick the better video; however, all current automatic evaluation systems lag behind human performance from most perspectives, revealing critical weaknesses in T2LV evaluation.
    }
  \label{fig:overview}
\end{figure}

To this end, we propose the synthetic long-video meta-evaluation (\textit{SLVMEval}) benchmark for T2LV.
The proposed benchmark is designed to meta-evaluate how effectively evaluation systems satisfy the minimum requirements for assessing long videos, specifically whether they can produce evaluations that are similar to those made by humans.
An overview of the proposed SLVMEval benchmark is shown in Figure~\ref{fig:overview}.
The SLVMEval benchmark comprises pairs of high-quality and low-quality videos.
Here, the high-quality videos are sampled from existing dense video-captioning datasets, and the low-quality videos are synthesized by degrading them with respect to specific aspects and then filtered manually.
The resulting videos range from several minutes to several hours, which is well beyond the length of video generated by current T2V models.
In this study, we employed the proposed SLVMEval benchmark to evaluate several baseline automatic evaluation systems, e.g., VLM-as-a-judge and CLIPScore.
Furthermore, in human evaluations, crowdworkers distinguished the high-quality video from the low-quality video at 84.7\%--96.8\% accuracy.
In contrast, existing evaluation systems lag human evaluators in nine of the ten aspects.
This gap is particularly large for evaluation aspects requiring semantic and temporal consistency between the video and the prompt.
In addition, for nearly all automatic evaluation systems, we observed a decrease in accuracy as the duration of the video increased, which suggests that existing evaluation systems are limited when evaluating long videos.

When we compared the evaluation results on the SLVMEval benchmark obtained with and without applying crowdworker-based filtering to the test data, we observed a strong correlation between them.
This suggests that our methodology can be employed to construct and extend the benchmark without always relying on expensive manual data filtering processes.

\section{Related work}
\label{sec:related_work}

\paragraph{Text-to-video generation.}
Recently, research on T2V models has attracted increasing attention.
Generally, T2V models can be categorized into two main approaches, i.e., those based on diffusion models~\cite{HaCohen2024LTXVideo,jin2025pyramidal} and those based on autoregressive transformers~\cite{deng2025autoregressive}.
Among the diffusion-based T2V models, Sora~\cite{openai2024videosim} supports video generation up to a duration of 60 s, and Kling~\cite{kling} supports up to 2 min.
In addition, by constructing a pipeline that generates videos autoregressively in chunk-based units, StreamingT2V~\cite{Henschel_2025_CVPR} enables the generation of highly consistent videos exceeding 2 min.
Phenaki~\cite{villegas2023phenaki} is an autoregressive transformer-based T2V model that takes discretized video tokens as input and outputs them in an autoregressive manner. 
By predicting the next video token, Phenaki is theoretically capable of generating videos of arbitrary lengths. 
However, only 1.4-s videos are used to train this model, which highlights that generating coherent, long videos remains a significant challenge.
Our framework aims to construct an environment that evaluates automatic evaluation systems in advance, which will be required to develop effective T2LV models in the future.

\paragraph{Meta-evaluation of T2V evaluation systems.}
Numerous benchmarks have been proposed to evaluate T2V models~\cite{liu2023fetv,Liu_2024_CVPR,huang2023vbench,liu2025uvemllmsunifiedevaluators}.
For example, VBench~\cite{huang2023vbench} provides fine-grained human annotations along dimensions, e.g., object fidelity, aesthetics, and technical quality. Its successors, VBench++~\cite{huang2024vbenchcomprehensiveversatilebenchmark} and VBench2~\cite{zheng2025vbench20advancingvideogeneration}, extend this coverage to additional aspects and families of models.
In contrast, UVE-Bench~\cite{liu2025uvemllmsunifiedevaluators} focuses on language model–based evaluation, offering human pairwise comparison data to evaluate the alignment between automatic judges and human preference.
Despite their merits, these methods are limited to clips of approximately 10 s.
To address this limitation, we construct a meta-evaluation testbed for long videos, i.e., up to several hours, which enables the development of automatic metrics that are suitable for future T2V systems.

\paragraph{Long-video understanding with VLMs.}
One goal of VLMs is to understand long-video content that humans watch daily, and their ability to process such videos is advancing~\cite{wang2024lvbenchextremelongvideo}.
For example, InternVideo2.5~\cite{chen2024fargpt4vclosinggap} can process videos of several minutes to hours.
The long-video understanding capabilities of these VLMs are predominantly evaluated through visual question answering tasks~\cite{Zhou_2025_CVPR,wang2024lvbenchextremelongvideo,NEURIPS2024_329ad516,nagrani2025neptunelongorbitbenchmarking,Tapaswi_2016_CVPR}.
The MLVU~\cite{Zhou_2025_CVPR} benchmark evaluates content understanding abilities on long videos, including movies, surveillance footage, and first-person view videos, through a wide range of tasks, including inference, anomaly detection, and summarization.
This benchmark has demonstrated that VLM performance on long videos deteriorates significantly as the duration increases. For example, the performance of Video-LLaMA-2~\cite{zhang-etal-2023-video} is reduced to chance level when videos are extended from 3 to 10 min.
However, these long-video benchmarks focus on content understanding rather than meta-evaluating automatic evaluation systems for T2V generation. Furthermore, they do not evaluate whether such systems can detect controlled quality degradations that are easily recognized by humans.
In contrast, the proposed SLVMEval benchmark utilizes controlled "high-quality versus low-quality" pairs to assess whether automatic evaluation systems for T2V models can reliably judge both \textit{video quality} and \textit{video-text consistency} on long videos.

\section{Meta-evaluation framework}
\label{sec:meta_evaluation_framework}
In this study, we adopt a pairwise comparison approach for the meta-evaluation framework, following both VBench and UVE~\cite{huang2023vbench,liu2025uvemllmsunifiedevaluators}.

\subsection{Evaluation system}
\label{subsec:evaluation_system}
Here, $\mathcal{P}$ and $\mathcal{V}$ denote the set of textual prompts and the set of all possible videos, respectively.
Then, we consider a T2V model as a mapping $g: \mathcal{P} \rightarrow \mathcal{V}$.
For any prompt $p \in \mathcal{P}$, an evaluation system $e$ receives the pair of videos $\{u_{p}, v_{p}\} = \{g(p), g'(p)\}$ produced by two T2V models $g$ and $g'$, and the system returns the video it determines to be of higher quality, i.e.,\ $e(p, \{u_{p}, v_{p}\})=z$, where $z \in \{u_{p}, v_{p}\}$.

\subsection{Meta-evaluation criteria}
\label{subsec:meta_evaluation_criteria}
As a metric for pairwise comparison, we measure an evaluation system's reliability with accuracy, the proportion of video pairs for which the system selects the better video.
To compute this metric, we build a test set
$\mathcal{D}= \{(p_i,\{v_{p_i}^{+},\,v_{p_i}^{-}\})\}_{i=1}^{N}$,
where each prompt $p_i$ has two generated videos:
$v_{p_i}^{+}$ (higher quality) and $v_{p_i}^{-}$ (lower quality).  
\begin{equation}
\hspace{-2mm}
       \mathrm{acc}(e,\mathcal{D}) = 
       \frac{1}{|\mathcal{D}|}
      \sum_{(p,\{v_{p}^{+},v_{p}^{-}\}) \in \mathcal{D}}
    \!\!\!\!\!\!
      \mathbf{1}\!\bigl[e(p,\{v_{p}^{+},v_{p}^{-}\}) = v_{p}^{+}\bigr],
\end{equation}
where $\mathbf{1}[\cdot]$ is the indicator function that returns $1$ when the evaluation system  selects $v_{p}^{+}$ and $0$ otherwise.
A higher accuracy indicates that the evaluation system is better at judging video quality from a particular aspect.

\section{Dataset construction}
\label{sec:dataset_construction}
The goal of this study is to meta-evaluate whether existing evaluation systems satisfy the minimum requirements to correctly assess the quality of generated videos.
To this end, we construct synthetic video pairs that differ in terms of only one specified quality aspect. The synthetic pairs are then used to build a test set for fine-grained pairwise evaluation.

\subsection{Dataset definition}
\label{subsec:dataset_definition}
We construct a synthetic test set $\mathcal{D}$ that probes whether an evaluation system can correctly assess video quality relative to several fundamental aspects.
Beginning with a manually curated prompt-video corpus $\mathcal{D}_{\mathrm{src}}=\{(p_i, v_{p_i}^{+})\}_{i=1}^{N}$, for each aspect $a$, we generate a degraded counterpart for each original video and construct the corresponding aspect-specific dataset $\mathcal{D}_{a}$:

\begin{equation}
    \begin{split}
      & \mathcal{D}_{a} = \{(p, \{v_{p}^{+}, v_{p}^{-}\}) \mid (p, v_{p}^{+}) \in \mathcal{D}_{\mathrm{src}}\}, \\
      & \text{where}\quad v_{p}^{-}=\Phi_{a}^{\text{low}}(p, v_{p}^{+}) .
    \end{split}
\end{equation}

Here, the function $\Phi_{a}^{\text{low}}$ takes a pair $(p, v_{p}^{+})$ and returns a degraded video $v_{p}^{-} \in \mathcal{V}$ for the aspect $a$.
Note that we apply a different function and construct a separate test set for each aspect.
Additional details about the degrading operation are presented in Section~\ref{subsec:degradation_operations}.

\subsection{Dataset source}
\label{subsec:dataset_source}
In this study, we selected the Vript dataset~\cite{NEURIPS2024_6903a5aa} as the source dataset.
We selected the Vript dataset because it comprises humanmade long videos that align with the purpose of our study.
In addition, the Vript dataset is designed for the dense video-captioning task~\cite{DBLP:conf/iccv/KrishnaHRFN17}; thus, each video in the dataset is segmented into semantically coherent clips and annotated with descriptive captions.
For evaluation, we sample each video at 1 fps and resize each frame such that its longer side measures 512 pixels.

Here, $v\in\mathcal{V}$ denotes a video with $T$ frames, and $f_{v,t}$ denotes the $t$-th frame of the video.
If the video is split into $M$ clips, we represent the $m$-th clip as $c_{v,m}=(f_{v,t})_{t\in[t_m,t_{m+1})}$, where $t_m$ and $t_{m+1}$ mark the first frames of the $m$-th and $(m+1)$-th clips, respectively.
We then assign a caption $p_m$ to each clip $c_{v,m}$ and define the prompt for $v$ as the concatenation $p=p_1\oplus p_2\oplus\cdots\oplus p_M$.

\begin{figure*}[t]
   \centering
   \adjincludegraphics[
     width=0.96\textwidth,
     clip,
     trim={{.0\width} {.70\height} {.0\width} {0}}
   ]{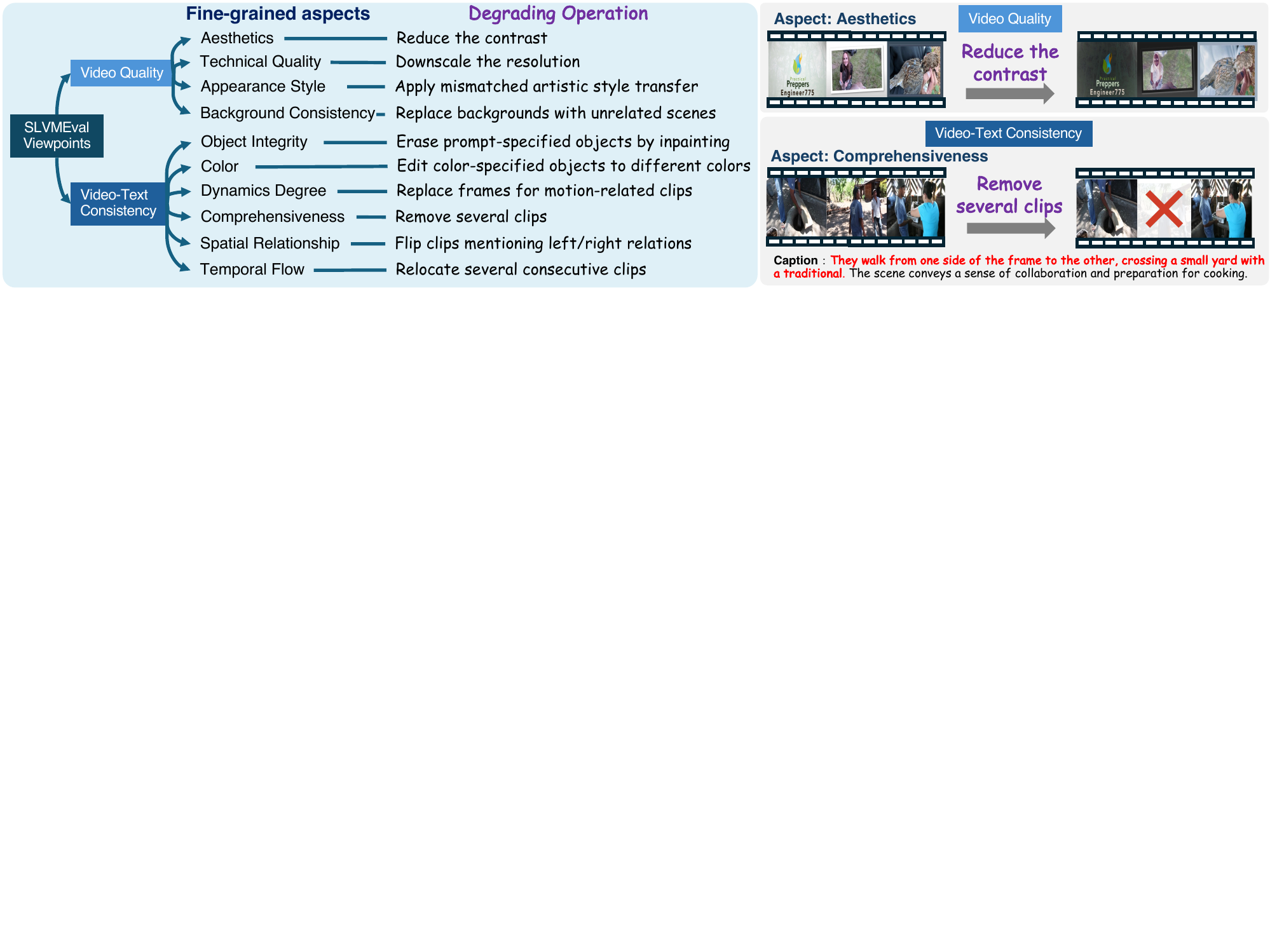}
  \caption{
  Viewpoints and aspect-specific degrading operations in the proposed SLVMEval benchmark. We organize the benchmark into two groups, i.e., \textit{video quality} and \textit{video-text consistency}, and define 10 aspects. For each aspect, we construct paired videos by applying a controlled synthetic degradation to the original long video while keeping all other factors unchanged. The right panels show example pairs. These controlled pairs enable precise meta-evaluation of whether an automatic evaluation system can reliably identify the high-quality video under each viewpoint. Additional example pairs are provided in the supplementary material.
  }
  \label{fig:example_videos}
\end{figure*}

\begin{algorithm}[t]
  \caption{Overview of $\Phi_{a}^{\text{low}}$}
  \label{alg:phi_low_overview}
  \begin{algorithmic}[1]
    \Require Prompt $p \in \mathcal{P}$, Video $v_{p} \in \mathcal{V}$
    \Ensure $v_{p}^{-}$
    \State $S \gets \textsc{SampleIdx}(M_{v_{p}},\,5)$  \Comment{random sample 5 clips}
    \State $F \gets [\,]$
    \For{$m \gets 0$ \textbf{to} $M_{v_{p}}-1$}
      \State $clip \gets c_{v_{p},m}$
      \If{$m \in S$}
        \State $F \gets F \cup \textsc{DegradeClip}(clip,\,p_{m})$
      \Else
        \State $F \gets F \cup clip$
      \EndIf
    \EndFor
    \State $v_{p}^{-} \gets \textsc{Concat}(F)$
    \State \Return $v_{p}^{-}$
  \end{algorithmic}
\end{algorithm}

\subsection{Degrading operations}
\label{subsec:degradation_operations}
Following prior work, we define multiple evaluation aspects.
However, systems to generate long videos are in their early stages; thus, it is difficult to identify these aspects solely from the failure modes observed in such videos.
Therefore, using benchmarks for short videos as references, we define 10 evaluation aspects grouped into the \textbf{video quality} and \textbf{video-text consistency} categories.
Using these aspects, we meta-evaluate whether an evaluation system possesses at least the minimal capabilities required for effective T2LV evaluation.

This categorization is inspired by comprehensive benchmarks, e.g., VBench~\cite{huang2023vbench}; however, in this study, we also adapted aspects of long-range temporal alignment, e.g., temporal flow and comprehensiveness, to probe the specific challenges associated with T2LV generation.
For each aspect, we independently invoke \textsc{DegradeClip} (Algorithm~\ref{alg:phi_low_overview}, line 6) to create a degraded video $v_{p}^{-}$.
Examples of the degraded outputs are shown in Figure~\ref{fig:example_videos}.
In the following, we introduce each aspect and explain the associated degrading operation.\footnote{See the supplementary material for additional implementation details.}

\subsubsection{Video Quality}
This category includes four aspects, i.e., aesthetics, technical quality, appearance style, and background consistency, which are used to evaluate the visual appearance.

\noindent\textbf{Aesthetics.}
We assess whether the evaluation system can judge frame-level aesthetic quality~\cite{huang2023vbench}.
Here, we reduce the contrast of the selected clips in the original video to create the degraded video $v_{p}^{-}$. \\
\textbf{Technical quality.}
We assess whether the evaluation system can detect low-resolution artifacts~\cite{liu2025uvemllmsunifiedevaluators}.
Here, we downscale the resolution of the selected clips from the original video to create $v_{p}^{-}$. \\
\textbf{Appearance style.}
We assess whether the evaluation system can detect an unnatural video style.
Here, to create $v_{p}^{-}$, we convert all frames in the selected clips to a different style (e.g., oil painting, manga, watercolor, or sketch) using the style-transfer API in OpenCV~\cite{opencv_library}. \\
\textbf{Background Consistency.}
We assess whether the evaluation system can determine if the background remains visually stable across frames.
Here, to create $v_{p}^{-}$, we remove the original backgrounds using rembg~\cite{rembg} and replace them with randomly sampled landscape images from the nature-dataset~\cite{nature_dataset_hf}.

\subsubsection{Video-text consistency}
This category covers aspects that are dependent on video semantics.
Here, we consider six aspects, i.e., temporal flow, comprehensiveness, object integrity, spatial relationship, dynamics degree, and color, which test whether the evaluation system can assess consistency between the video and the corresponding text prompt.\\
\textbf{Temporal flow.}
We assess whether the evaluation system can follow the order of events in a video.
We create $v_{p}^{-}$ by moving five consecutive clips to random positions, thereby breaking the temporal order of events in the original video. \\
\textbf{Comprehensiveness.}
We assess whether the evaluation system can verify that a video fully covers the prompt.
Here, we randomly remove five clips from the original video to create $v_{p}^{-}$. \\
\textbf{Object integrity.}
We assess whether the evaluation system can recognize the presence of prompt-specified objects.
To create $v_{p}^{-}$, we first extract object names from each prompt $p_m$ using Qwen3-8B~\cite{qwen3technicalreport} and localize them using Grounding~DINO~\cite{10.1007/978-3-031-72970-6_3}. Finally, we erase them via Stable-Diffusion-Inpainting~\cite{Rombach_2022_CVPR}. \\
\textbf{Spatial relationship.}
We assess whether the evaluation system can determine if the spatial relations specified in the prompt (left/right) match each frame.
Here, to create $v_{p}^{-}$, we first employ Qwen3-8B to identify clips whose captions mention left/right relations for some object, and then we horizontally flip all frames in the identified clips. \\
\textbf{Dynamics degree.}
We assess whether the evaluation system can determine if the amount of motion specified in the target prompt is appropriate.
To create $v_{p}^{-}$, we first employ Qwen3-8B to identify clips whose captions mention object motion, and then we replace each frame in the identified clips with the middle frame of the clip, effectively producing a static clip without motion. \\
\textbf{Color.}
We assess whether the evaluation system can judge if the colors specified in the prompt match the content of the generated video.
For this assessment, we employ Qwen-Image-Edit-2509~\cite{wu2025qwenimagetechnicalreport} to modify the colors of objects in all frames of the relevant clips to create $v_{p}^{-}$. We use Qwen3-8B to identify which clips mention an object’s color. %

\begin{table*}[t]
\caption{Statistics of existing and proposed benchmarks. VBench-Long~\cite{huang2024vbenchcomprehensiveversatilebenchmark} only provides a framework for automatic evaluation (it does not include human annotations of video quality). It supplies prompts as inputs to T2V models, assuming generated videos of approximately 1 min~\cite{Huang2024VBenchSlides}.}
  \centering
  \small
    \begin{tabular}{lccrrrr}
\toprule
 & $\!$Human Annotation$\!$ & $\!$Aspects$\!$ & $\!$Videos$\!$ & $\!$Unique Prompts$\!$ & $\!$Max Duration (sec)$\!$ & $\!$ Avg. Prompt Length (chars) \\
\midrule
UVE-Bench~\cite{liu2025uvemllmsunifiedevaluators} & $\checkmark$ & 15 & 1{,}045 & 293 & 6.1 & 73.68 \\
VBench~\cite{huang2023vbench} & $\checkmark$ & 16 & 21{,}110 & 968 & 3.3 & 41.32 \\
VBench Long~\cite{huang2024vbenchcomprehensiveversatilebenchmark} &  & 16 & N/A & 944 & N/A & 41.00 \\
SLVMEval (\textbf{ours}) & $\checkmark$ & 10 & 3{,}932 & 1{,}461 & 10{,}486.0 & 57{,}883.52 \\
\bottomrule
\end{tabular}

  \label{tab:benchmark_comp_stats}
\end{table*}

\subsection{Human-Annotated Dataset Filtering}
\label{subsec:human_annotation}
To ensure the quality and validity of the synthesized degraded video pairs, we apply human annotations and filter the dataset accordingly.
The goal of this process is to verify that each degradation is applied as intended and is clearly perceptible to human observers.

\paragraph{Annotation Rules.}
In this study, we showed the crowdworkers the target aspect, the paired videos $\{v_{p}^{+}, v_{p}^{-}\}$, and a description of the applied degradation.
For each pair, five crowdworkers rated the success of the degradation on a three-point A/B/C scale.
\begin{itemize}[leftmargin=1.5em]
    \item[\textbf{A.}] The degradation clearly succeeds in all selected clips.
    \item[\textbf{B.}] The degradation succeeds in at least one but not all selected clips, or the effect is weak.
    \item[\textbf{C.}] The degradation fails in all selected clips.
\end{itemize}

\paragraph{Dataset filtering.}
Based on the ratings provided by the five crowdworkers, we selected samples to include in the final test set (i.e., those deemed to have been degraded successfully).
Here, we used the following success conditions.
\begin{enumerate}
    \item \textbf{No C votes.} We excluded the samples for which any crowdworker judged the degradation a complete failure.
    \item \textbf{Votes for A exceed votes for B.} We retained only the samples for which the crowdworker agreement leaned toward “clear success” rather than “partial success.”
\end{enumerate}
This filtering process ensured that each sample exhibited degradation that humans could clearly agree on.\footnote{
The supplementary material details the annotation platform (Yahoo!~Crowdsourcing), compensation, and quality control.
}

\subsection{Dataset statistics}
For each evaluation aspect, we binned all candidate videos by duration in 100-s increments and randomly sampled up to 20 videos from each bin.
Then, we performed manual filtering and retained only the videos in which the degradation worked as intended.
In total, 3{,}932 videos remained.
The resulting clips averaged 1141 s ($\simeq$ 19 min) in length, with the longest being 10,486 s ($\simeq$ 2 h and 54 min).
Furthermore, we confirmed that the dataset spanned 15 distinct content categories in a balanced manner.
\footnote{
The detailed distributions for both video duration and content categories are presented in the supplementary material.
}

Table~\ref{tab:benchmark_comp_stats} compares the dataset constructed in this study with existing meta-evaluation datasets.
Compared with manually annotated resources, e.g., VBench~\cite{huang2023vbench} and UVE~\cite{liu2025uvemllmsunifiedevaluators}, our dataset features significantly longer prompts and considerably longer videos, providing a more challenging testbed for the evaluation of long videos.

\section{Baseline automatic evaluation systems}
\label{sec:baseline_systems}

\subsection{Pairwise VLM-as-a-judge}
\label{subsec:pairwise_VLM_as_a_judge}
Recently, several studies have proposed using VLMs as evaluation systems~\cite{lee-etal-2024-prometheus,liu2025videolanguagemodelreliable}.
To assess their effectiveness for T2LV evaluation, we implemented two variants of VLM-as-a-judge, i.e., a \textbf{video-based evaluation} variant and a \textbf{text-based evaluation} variant.
For both systems, we employed the GPT-5 and GPT-5-mini~\cite{OpenAI2025GPT5SystemCard} as proprietary models, and we employed the Qwen3-VL-235B~\cite{qwen3technicalreport,Qwen2.5-VL,Qwen2VL,Qwen-VL} as an open-weight model.

\paragraph{Video-based evaluation.}
\label{para:video_based_evaluation}
In the video-based evaluation, given a prompt $p$ and two candidate videos $u_{p}, v_{p} \in \mathcal{V}$, the VLM predicts which video is better in terms of aspect $a$.
Formally, this process is performed as follows:
\begin{equation}
    \!e_{a}(p, \!\{u_{p},\! v_{p}\}) \!\!=\!\! \begin{cases}
      u_{p}, & \!\!\!\!\!\text{if } \textsc{VLM}(a, p, u_{p}, v_{p}) \!=\!\! \text{``first''}, \\
      v_{p}, & \!\!\!\!\!\text{if } \textsc{VLM}(a, p, u_{p}, v_{p}) \!=\!\! \text{``second''}.\!\!\!
    \end{cases}
\end{equation}

Here, $\textsc{VLM}$ takes the prompt $p$, the videos $u_{p}$ and $v_{p}$, and the evaluation aspect $a$ as input, and it outputs either “first” or “second” to indicate which video is preferred.

\paragraph{Text-based evaluation.}
\label{para:text_based_evaluation}
In the text-based evaluation, the VLM first generates captions $caption_{u_{p}}$ and $caption_{v_{p}}$ for $u_{p}$ and $v_{p}$, respectively.
Then, a language model (LM) compares these captions with the prompt $p$ and selects the caption that aligns more closely with the prompt. The corresponding video is then returned as the final output.
Formally, this process is performed as follows:
\begin{equation}
    \begin{split}
        & d = \textsc{LM}(a, p, \textsc{VLM}_{\mathrm{cap}}(a, u_{p}), \textsc{VLM}_{\mathrm{cap}}(a, v_{p})), \\
        & e_{a}(p, \{u_{p}, v_{p}\}) = \begin{cases}
          u_{p}, & \text{if } d = \text{``first''}, \\
          v_{p}, & \text{if } d = \text{``second''}.
        \end{cases}
    \end{split}    
\end{equation}

Here, $\textsc{VLM}_{\mathrm{cap}}$ is a video-captioning model that generates a caption for the input video relative to aspect $a$.
$\textsc{LM}$ takes the prompt $p$ and the generated captions $caption_{u_{p}}$ and $caption_{v_{p}}$ as input, and it returns a decision $d \in \{\text{“first” }, \text{“second” }\}$ indicating which caption better matches the prompt.
For $\mathrm{VLM}_{\mathrm{cap}}$, we employed the GPT-5-mini and Qwen3-VL-235B in the closed and open-weight settings, respectively.
Note that this approach can be considered a form of roundtrip evaluation~\cite{somers-2005-round}, and similar metrics are employed in BLIP-BLEU and the EvalCrafter baseline system~\cite{Liu_2024_CVPR}.
We expect that projecting a video into text space can make aspects that are difficult to evaluate from raw frame sequences easier to assess.

For the GPT-5 family, context-length limits prevented us from inputting all frames to the model.
Thus, we automatically detected clip boundaries using FFmpeg~\cite{tomar2006converting} and extracted the center frame from each clip.
These frames were then input to the VLM as a concise summary of the video.\footnote{The source dataset also includes clip-boundary annotations.
However, we did not use these annotations in the current study because ground truth boundaries are not available in practical T2V model evaluation scenarios.}
Furthermore, to mitigate LM-specific biases, e.g., position bias~\cite{Tian_2025_CVPR}, we evaluated each pair twice by swapping the order of $u_{p}$ and $v_{p}$.\footnote{The evaluation prompts are listed in the supplementary material.}

\subsection{CLIPScore}
\label{subsec:clip_score}
In this study, we used CLIPScore~\cite{hessel-etal-2021-clipscore} as an existing evaluation system
CLIPScore is an automatic evaluation system based on the similarity between text and visual content, and it is widely used in T2V tasks~\cite{singer2023makeavideo,Sun_2025_CVPR}.

The CLIPScore-based system is defined as follows in our framework:
\begin{align}
     e\bigl(p,\{u_{p}, v_{p}\}\bigr)
    &=  \operatorname*{arg\,max}_{s \in \{u_{p}, v_{p}\}} \frac{1}{N_{s}^{\prime}}\sum_{i=1}^{N_{s}^{\prime}} \operatorname{CLIPScore}\left(f_{s,i}^{\mathrm{mid}}, p\right) \nonumber\\
     \text{where }\quad
     f_{s,i}^{\mathrm{mid}} &= f_{s,{\lfloor\frac{t_{i} + t_{i + 1}}{2}}\rfloor}
    .
\end{align}
Here, $N_{s}^{\prime}$ denotes the number of clips detected in video $s$.
We employed FFmpeg to segment the videos (Section~\ref{subsec:pairwise_VLM_as_a_judge}).
Here, $\operatorname{CLIPScore}\left(f_{s,i}^{\mathrm{mid}}, p\right)$ is the CLIPScore between the center frame of the $i$-th detected clip and the prompt $p$.
We employed Jina CLIP v2~\cite{DBLP:journals/corr/abs-2412-08802} as the base model because it can handle text inputs of up to 8192 tokens.

\subsection{VideoScore}
\label{subsec:videoscore}
We used VideoScore-v1.1~\cite{he-etal-2024-videoscore} as one of our baseline evaluation systems.

VideoScore extends a pretrained VLM by adding a regression head, and it is further fine-tuned on the VideoFeedback dataset ~\cite{he-etal-2024-videoscore}, which contains human video quality ratings.
VideoScore can assign a quality score to a video from five aspects.
Note that the aspects defined in this study differ from those defined by VideoScore; thus, we established a mapping between our aspects and those in VideoScore.\footnote{The mapping rules are defined in the supplementary material.
}
In our framework, we define the VideoScore-based system as follows:
\begin{equation}
  e_{a}(p, \{u_{p}, v_{p}\}) = \operatorname*{arg\,max}_{s \in \{u_{p}, v_{p}\}} \mathrm{VideoScore}_{a}(s, p).
\end{equation}
Here, $\mathrm{VideoScore}_{a}(s, p)$ denotes the VideoScore for video $s$ and prompt $p$ under aspect~$a$.
VideoScore can handle at most 48 frames; thus, as in experiments with the the GPT-5 family, we provided the central frame of each scene split by FFmpeg as a representative frame to VideoScore.
Here, if the sequence length was still exceeded, we randomly sampled up to 48 frames from each video.

\section{Experiments}
\label{sec:experiments}
Our experiment aims to evaluate the robustness of current evaluation systems when they are applied to long videos.

\subsection{Experimental setup}
\label{subsec:experimental_setup}
Here, we evaluated the eight evaluation systems described in Section~\ref{sec:baseline_systems}, i.e., pairwise VLM-as-a-judge (video-based and text-based), CLIPScore, and VideoScore.
Furthermore, we used accuracy (defined in Section~\ref{subsec:meta_evaluation_criteria}) as the meta-evaluation metric. Here, accuracy was calculated as the proportion of video pairs where a system correctly identified the original video as having higher quality.

As a comparison to automatic evaluation systems, we also measured the performance of humans tasked with judging video quality.
Here, using the same interface employed for the evaluation systems, each worker was shown a pair of videos and the corresponding prompt, and they were asked which video they considered to be higher quality. Then, we computed their accuracy in the same manner.\footnote{Additional details about the experimental setup (e.g., human evaluation processes) are described in the supplementary material.}

\subsection{Results and discussion}
\label{subsec:experimental_results}

In this section, we discuss the observations and insights we draw from our comprehensive evaluation experiments.

\newcolumntype{C}[1]{>{\centering\arraybackslash}m{#1}}
\renewcommand\cellalign{cc} %

\newcommand{\AspectFontSizePt}{7}   %
\newcommand{\AspectBaseLinePt}{8}   %

\renewcommand\theadfont{%
  \fontsize{\AspectFontSizePt}{\AspectBaseLinePt}\selectfont\bfseries
}
\renewcommand\theadgape{} %

\newcommand{\CIFontSizePt}{6.0}      %
\newcommand{\CIFontBaseLinePt}{7.2}  %

\newcommand{\cifont}{%
  \fontsize{\CIFontSizePt}{\CIFontBaseLinePt}\selectfont
}

\newcommand{\accpm}[2]{#1{\cifont$\pm$#2}}

\newcommand{\CategoryFontSizePt}{8}
\newcommand{\CategoryBaseLinePt}{9}

\newcommand{\categoryfont}{%
  \fontsize{\CategoryFontSizePt}{\CategoryBaseLinePt}\selectfont\bfseries
}

\definecolor{secondgreen}{HTML}{00a760}

\begin{table*}[t]
\caption{
  Accuracy (\%) of each baseline system on SLVMEval.
  Numbers are accuracy\,\% $\pm$ 95\% CI (percentage points).
  \textcolor{blue}{Blue} bold = best per aspect; \textcolor{secondgreen}{green} = second best.
  The chance level is 50\%.
}
  \centering
  \small %
  \setlength{\tabcolsep}{1.0pt}
  \renewcommand{\arraystretch}{0.9}
  \begin{tabular*}{\textwidth}{@{\extracolsep{\fill}}l *{10}{C{0.08\textwidth}}}
    \toprule
    & \multicolumn{4}{c}{{\categoryfont Video Quality}}
    & \multicolumn{6}{c}{{\categoryfont Video-Text Consistency}} \\
    \cmidrule(lr){2-5}\cmidrule(lr){6-11}
    \noalign{\vspace{-1.5pt}}
    &
    \thead{Aesthetics} &
    \thead{Technical\\Quality} &
    \thead{Appearance\\Style} &
    \thead{Background\\Consistency} & %
    \thead{Object\\Integrity} &      %
    \thead{Color} &
    \thead{Dynamics\\Degree} &
    \thead{Comprehen-\\siveness} &
    \thead{Spatial\\Relationship} &
    \thead{Temporal\\Flow} \\[-1.5pt]
    \midrule

    Video-based \\
    \quad GPT-5
      & \accpm{\textcolor{secondgreen}{90.1}}{2.5}
      & \accpm{\textcolor{secondgreen}{85.8}}{4.2}
      & \accpm{\textcolor{secondgreen}{88.9}}{2.5}
      & \accpm{\textcolor{blue}{\textbf{98.9}}}{0.8} %
      & \accpm{72.0}{6.2}                            %
      & \accpm{\textcolor{secondgreen}{84.3}}{3.5}
      & \accpm{35.3}{3.6}
      & \accpm{51.3}{4.5}
      & \accpm{\textcolor{secondgreen}{59.7}}{4.4}
      & \accpm{50.3}{4.1} \\
    \quad GPT-5-mini
      & \accpm{84.0}{3.0}
      & \accpm{48.1}{6.1}
      & \accpm{78.0}{3.2}
      & \accpm{\textcolor{secondgreen}{95.2}}{1.6} %
      & \accpm{66.5}{6.5}                          %
      & \accpm{69.4}{4.5}
      & \accpm{31.5}{3.5}
      & \accpm{45.7}{4.5}
      & \accpm{51.1}{4.5}
      & \accpm{43.7}{4.1} \\
    \quad Qwen3
      & \accpm{55.7}{4.1}
      & \accpm{51.9}{6.1}
      & \accpm{55.3}{3.9}
      & \accpm{49.7}{3.7} %
      & \accpm{38.5}{6.7} %
      & \accpm{48.8}{4.9}
      & \accpm{50.0}{3.8}
      & \accpm{52.6}{4.5}
      & \accpm{51.7}{4.5}
      & \accpm{50.2}{4.1} \\

    Text-based \\
    \quad GPT-5
      & \accpm{74.8}{3.6}
      & \accpm{46.2}{6.1}
      & \accpm{81.1}{3.1}
      & \accpm{83.8}{2.7} %
      & \accpm{68.0}{6.5} %
      & \accpm{68.9}{4.5}
      & \accpm{43.1}{3.8}
      & \accpm{50.6}{4.5}
      & \accpm{47.0}{4.5}
      & \accpm{43.5}{4.1} \\
    \quad GPT-5-mini
      & \accpm{75.0}{3.6}
      & \accpm{53.8}{6.1}
      & \accpm{79.6}{3.2}
      & \accpm{81.1}{2.9} %
      & \accpm{65.5}{6.6} %
      & \accpm{71.8}{4.4}
      & \accpm{43.8}{3.8}
      & \accpm{50.6}{4.5}
      & \accpm{51.1}{4.5}
      & \accpm{41.2}{4.0} \\
    \quad Qwen3
      & \accpm{51.6}{4.1}
      & \accpm{50.0}{6.1}
      & \accpm{72.4}{3.5}
      & \accpm{73.0}{3.3} %
      & \accpm{51.0}{6.9} %
      & \accpm{61.0}{4.7}
      & \accpm{\textcolor{secondgreen}{52.7}}{3.8}
      & \accpm{51.7}{4.5}
      & \accpm{50.2}{4.5}
      & \accpm{\textcolor{secondgreen}{55.6}}{4.1} \\

    CLIPScore
      & \accpm{56.4}{5.8}
      & \accpm{72.3}{7.7}
      & \accpm{53.2}{5.5}
      & \accpm{68.6}{4.8}                      %
      & \accpm{\textcolor{secondgreen}{76.0}}{8.4} %
      & \accpm{66.2}{6.5}
      & \accpm{51.7}{5.4}
      & \accpm{\textcolor{secondgreen}{57.4}}{6.3}
      & \accpm{55.1}{6.3}
      & \accpm{50.5}{5.8} \\
    VideoScore
      & \accpm{52.5}{5.8}
      & \accpm{33.8}{8.1}
      & \accpm{65.7}{5.3}
      & \accpm{71.2}{4.7} %
      & \accpm{66.0}{9.3} %
      & \accpm{33.8}{6.5}
      & \accpm{48.6}{5.4}
      & \accpm{34.5}{6.1}
      & \accpm{49.6}{6.4}
      & \accpm{46.3}{5.8} \\
    Human
      & \accpm{\textcolor{blue}{\textbf{96.5}}}{2.1}
      & \accpm{\textcolor{blue}{\textbf{91.8}}}{4.7}
      & \accpm{\textcolor{blue}{\textbf{95.2}}}{2.4}
      & \accpm{95.0}{2.3}                          %
      & \accpm{\textcolor{blue}{\textbf{86.6}}}{6.7} %
      & \accpm{\textcolor{blue}{\textbf{96.8}}}{2.4}
      & \accpm{\textcolor{blue}{\textbf{95.9}}}{2.1}
      & \accpm{\textcolor{blue}{\textbf{84.7}}}{4.6}
      & \accpm{\textcolor{blue}{\textbf{88.2}}}{4.1}
      & \accpm{\textcolor{blue}{\textbf{86.6}}}{4.0} \\
    \bottomrule
  \end{tabular*}
  \label{tab:main_results_full}
\end{table*}

\paragraph{Benchmark validity.}
As shown in Table~\ref{tab:main_results_full}, the human evaluators achieved 84.7\%--96.8\% accuracy across all 10 aspects, which indicates that the differences between “high-quality” and “low-quality” videos, as defined by SLVMEval, are easily perceptible to humans.
Thus, the SLVMEval benchmark has an appropriate difficulty to test our core question, i.e., whether an evaluation system satisfies the basic capabilities required to assess video quality.

\begin{figure*}[t]
   \centering
   \includegraphics[width=0.99\linewidth]{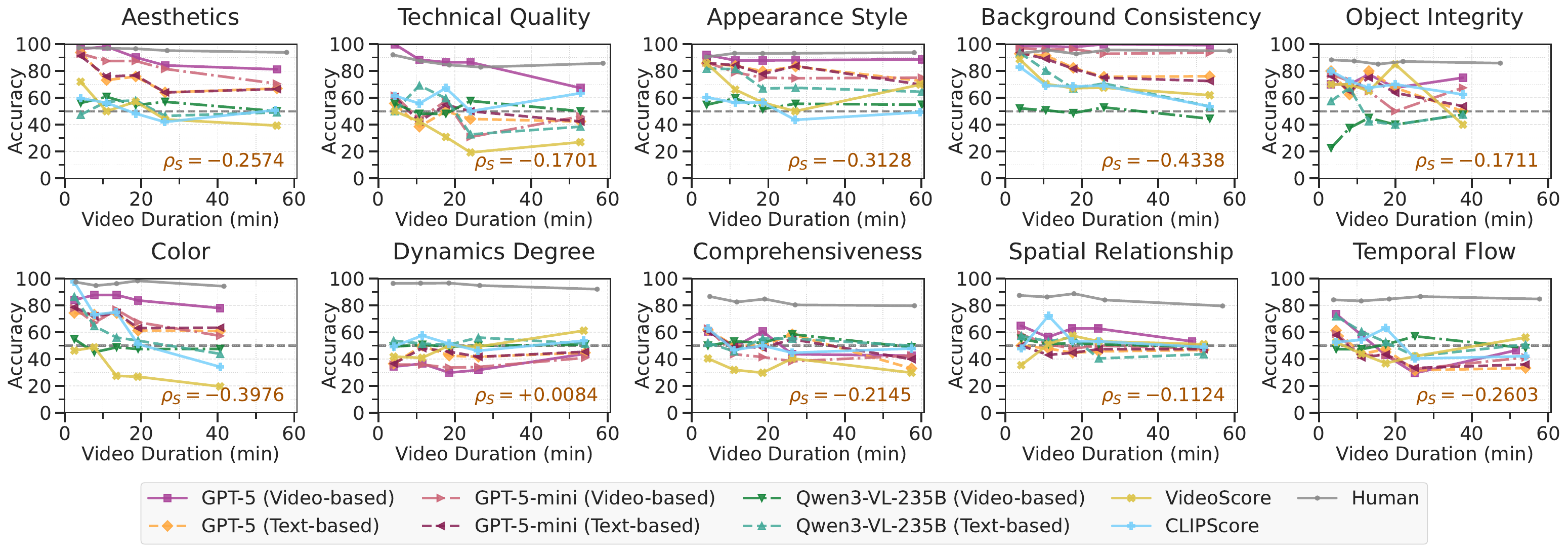}
   \caption{
        Relationship between video duration and accuracy. First, we sorted the dataset by video duration, divided it into four bins (intervals), and computed the accuracy within each bin. The x-axis and y-axis represents the average video duration in each bin and the corresponding accuracy, respectively. For each aspect and automatic evaluation system (excluding the human evaluators), we computed the Spearman rank correlation coefficient $\rho_{\mathrm{S}}$ between video duration and accuracy. Here, we sorted the samples by video duration, divided them into 50 bins, computed the accuracy within each bin, and then measured Spearman’s $\rho_{\mathrm{S}}$ using these 50 accuracy values. The per-aspect Spearman’s $\rho_{\mathrm{S}}$ values and associated $p$-values for each automatic evaluation system are reported in the supplementary material.
    }
   \label{fig:degrade_pos_vs_acc}
\end{figure*}

\begin{figure*}[t]
  \centering
  \includegraphics[width=0.95\linewidth]{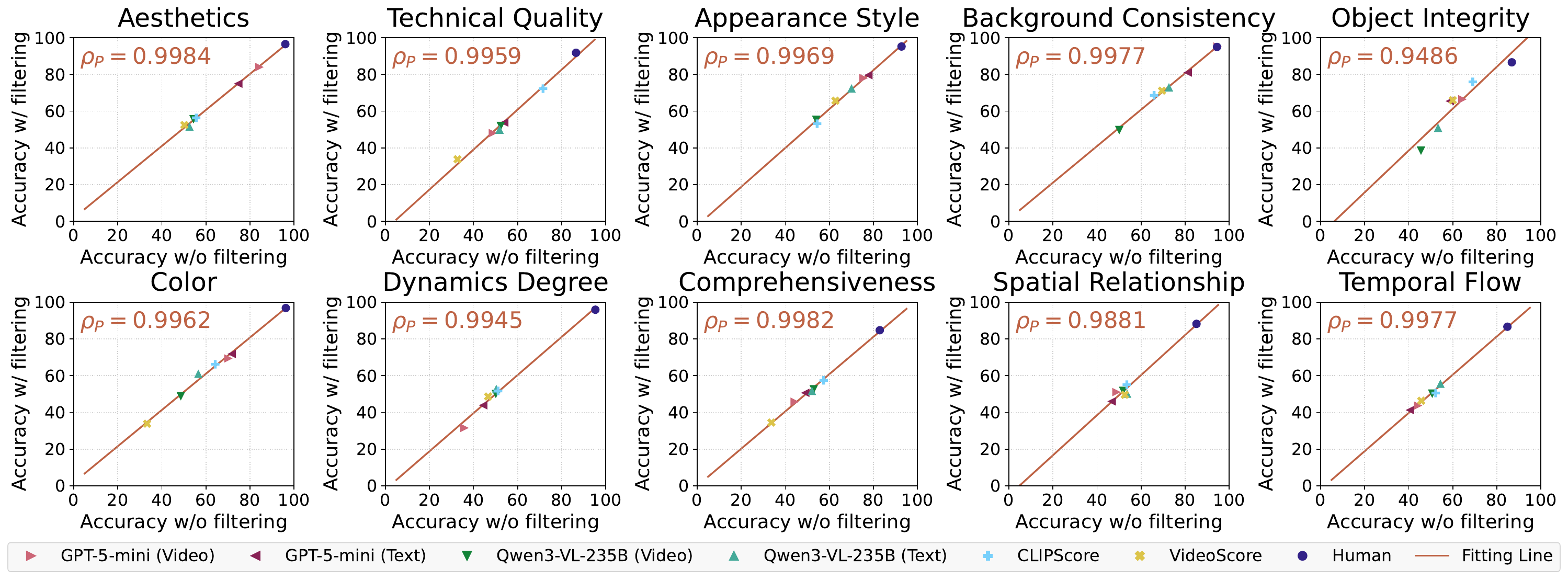}
  \caption{
    Relationship between accuracy values before and after filtering on the degraded SLVMEval data. For each aspect, we plot the accuracy of each evaluation system before versus after filtering and compute the Pearson correlation coefficient $\rho_{\mathrm{P}}$ from these points. Each marker corresponds to one evaluation system. The horizontal axis and vertical axis show the accuracy on the unfiltered data and the filtered test set, respectively. The solid line is the fitted linear regression line visualizing the correlation.
    }
    \label{fig:clean_vs_none_clean}
\end{figure*}

\paragraph{Overall trends.}
As shown in Table~\ref{tab:main_results_full}, the existing evaluation systems compared in this study lag behind human evaluators by 6.3\% and 43.2\% on nine of the 10 aspects.
Although these tasks are easy for humans, current systems struggle to assess long videos on many aspects.
Within the video quality category, GPT-5 (video-based evaluation) exceeded 85\% accuracy for all aspects, indicating a strong ability to detect surface-level degradations (e.g., frame-level visual issues or stylistic shifts); however, it still falls short of human performance.
In contrast, for the video-text consistency category, which requires both semantic alignment and temporal consistency between the prompt and the video content, the accuracy values were lower than in the Video Quality category.
These results suggest that even the strong proprietary models face difficulties with semantic alignment and temporal consistency in long videos.
For the text-based evaluation, Qwen3 obtained higher scores than in the video-based evaluation, with gains of 23.3 points in background consistency, 17.1 points in appearance style, and 12.5 points in object integrity.
These results suggest that text-based evaluation may be an effective option for improving evaluation performance for certain models and aspects.
In addition, the CLIPScore ranked second only to the human judgments in terms of object integrity and comprehensiveness.
We attribute this to CLIP’s contrastive pretraining, which yields strong text--image alignment and makes it especially sensitive when objects or clips explicitly mentioned in the prompt are missing.
In contrast, CLIPScore processes frames independently; thus, it remains at near chance level for aspects that require reasoning over adjacent frames, e.g., the dynamics degree and temporal flow aspects, which follows from the system’s design.
We found that VideoScore fell below the chance level for some aspects.
Here, we suspect this occurred because its five predefined criteria do not align with our aspects in a one-to-one manner, thereby leading to inconsistent judgments.

\paragraph{Sensitivity to video duration.}
Figure~\ref{fig:degrade_pos_vs_acc} shows how each system’s accuracy on SLVMEval varied with video duration for each aspect.
Here, we partitioned the dataset into subsets based on the video duration and computed the accuracy within each subset.
We then computed the Spearman rank correlation coefficient $\rho_{\mathrm{S}}$ between video duration and accuracy.
We found that, for all aspects (except dynamics degree), the accuracy values of most evaluation systems tended to decrease as the video duration increased.
Note that this trend was particularly evident for the background consistency and color aspects.
In contrast, for aspects on which all systems performed poorly (e.g., dynamics degree), the correlation between accuracy and video duration was relatively weak because the systems frequently failed even on short videos.
As a result, the $p$-values (averaged over aspects and systems) do not indicate statistical significance at the 0.05 level. Thus, these results should be interpreted as describing only a general tendency rather than a strong effect.\footnote{The correlation coefficients and $p$-values for all systems and aspects are shown in the supplementary material.
}

\paragraph{Necessity of data cleaning.}
Figure~\ref{fig:clean_vs_none_clean} shows the Pearson correlation coefficient $\rho_{\mathrm{P}}$ between the accuracy values obtained when evaluating each model on SLVMEval benchmark with and without manual data filtering.
These results suggest that, within the scope of our degradation operations, our procedure can produce reasonably reliable video pairs for evaluation even without costly manual filtering.
In turn, this implies that manual filtering is not strictly necessary for our degradation types when constructing domain-specific evaluation benchmarks or further extending the proposed benchmark, highlighting the scalability of our benchmark construction method built on the degrading operations.

\section{Conclusion}
\label{sec:conclusion}
This paper introduced the SLVMEval meta-evaluation benchmark to assess T2LV evaluation systems.
SLVMEval adopts a pairwise comparison framework to evaluate videos.
To create this benchmark, we begin with a dense video-captioning dataset and synthetically generate high-quality and low-quality video pairs.
During the meta-evaluation process, for each pair of videos and the specified aspect, the target evaluation system judges which video is higher quality.
As a baseline, we performed meta-evaluations using existing evaluation systems.
The results demonstrated that these systems underperformed humans in nine out of 10 aspects, despite the task being easy for humans.
In addition, the findings of this study suggest that the performance of evaluation systems tends to decline as video duration increases, pointing to the limitations of existing systems in assessing T2LV.
Furthermore, a comparison of the evaluation results on SLVMEval benchmark with and without manual filtering suggests that such filtering is not strictly required for our degradation operations.
In turn, this indicates that our approach can serve as a scalable method to expand the benchmark.
Finally, the focus of the proposed benchmark is to measure whether a system possesses the minimum capabilities required of an effective T2LV evaluation system.
We hope that future research on T2LV evaluation systems will consider achieving human-level performance on the proposed benchmark as an initial target.

\clearpage

\paragraph{Acknowledgments.}
This work was supported by JST Moonshot R\&D Grant Number JPMJMS2011-35 (fundamental research); the Ministry of Education, Culture, Sports, Science and Technology (MEXT) initiative “R\&D Hub Aimed at Ensuring Transparency and Reliability of Generative AI Models”; and JSPS KAKENHI Grant Number JP25KJ0615.
A part of the computation was carried out using the computer resource offered under the category of General Projects by Research Institute for Information Technology, Kyushu University.

{
    \small
    \bibliographystyle{ieeenat_fullname}
    \bibliography{main}

\begin{thebibliography}{52}
\providecommand{\natexlab}[1]{#1}
\providecommand{\url}[1]{\texttt{#1}}
\expandafter\ifx\csname urlstyle\endcsname\relax
  \providecommand{\doi}[1]{doi: #1}\else
  \providecommand{\doi}{doi: \begingroup \urlstyle{rm}\Url}\fi

\bibitem[Bai et~al.(2023)Bai, Bai, Yang, Wang, Tan, Wang, Lin, Zhou, and Zhou]{Qwen-VL}
Jinze Bai, Shuai Bai, Shusheng Yang, Shijie Wang, Sinan Tan, Peng Wang, Junyang Lin, Chang Zhou, and Jingren Zhou.
\newblock Qwen-vl: A versatile vision-language model for understanding, localization, text reading, and beyond.
\newblock \emph{arXiv preprint arXiv:2308.12966}, 2023.

\bibitem[Bai et~al.(2025)Bai, Chen, Liu, Wang, Ge, Song, Dang, Wang, Wang, Tang, Zhong, Zhu, Yang, Li, Wan, Wang, Ding, Fu, Xu, Ye, Zhang, Xie, Cheng, Zhang, Yang, Xu, and Lin]{Qwen2.5-VL}
Shuai Bai, Keqin Chen, Xuejing Liu, Jialin Wang, Wenbin Ge, Sibo Song, Kai Dang, Peng Wang, Shijie Wang, Jun Tang, Humen Zhong, Yuanzhi Zhu, Mingkun Yang, Zhaohai Li, Jianqiang Wan, Pengfei Wang, Wei Ding, Zheren Fu, Yiheng Xu, Jiabo Ye, Xi Zhang, Tianbao Xie, Zesen Cheng, Hang Zhang, Zhibo Yang, Haiyang Xu, and Junyang Lin.
\newblock Qwen2.5-vl technical report.
\newblock \emph{arXiv preprint arXiv:2502.13923}, 2025.

\bibitem[Bradski(2000)]{opencv_library}
G. Bradski.
\newblock {The OpenCV Library}.
\newblock \emph{Dr. Dobb's Journal of Software Tools}, 2000.

\bibitem[Brooks et~al.(2024)Brooks, Peebles, Holmes, DePue, Guo, Jing, Schnurr, Taylor, Luhman, Luhman, Ng, Wang, and Ramesh]{openai2024videosim}
Tim Brooks, Bill Peebles, Connor Holmes, Will DePue, Yufei Guo, Li Jing, David Schnurr, Joe Taylor, Troy Luhman, Eric Luhman, Clarence Ng, Ricky Wang, and Aditya Ramesh.
\newblock Video generation models as world simulators, 2024.

\bibitem[Chen et~al.(2025)Chen, Lin, Yang, Lin, Zhu, Fan, Zhang, Chen, Chen, Ma, Xiong, Wang, Pang, Kang, Xu, Jin, Liang, Song, Zhao, Xu, Qiu, Li, Fei, Li, and Zhou]{chen2025skyreelsv2infinitelengthfilmgenerative}
Guibin Chen, Dixuan Lin, Jiangping Yang, Chunze Lin, Junchen Zhu, Mingyuan Fan, Hao Zhang, Sheng Chen, Zheng Chen, Chengcheng Ma, Weiming Xiong, Wei Wang, Nuo Pang, Kang Kang, Zhiheng Xu, Yuzhe Jin, Yupeng Liang, Yubing Song, Peng Zhao, Boyuan Xu, Di Qiu, Debang Li, Zhengcong Fei, Yang Li, and Yahui Zhou.
\newblock Skyreels-v2: Infinite-length film generative model, 2025.

\bibitem[Chen et~al.(2024)Chen, Wang, Tian, Ye, Gao, Cui, Tong, Hu, Luo, Ma, Ma, Wang, Dong, Yan, Guo, He, Shi, Jin, Xu, Wang, Wei, Li, Zhang, Zhang, Cai, Wen, Yan, Dou, Lu, Zhu, Lu, Lin, Qiao, Dai, and Wang]{chen2024fargpt4vclosinggap}
Zhe Chen, Weiyun Wang, Hao Tian, Shenglong Ye, Zhangwei Gao, Erfei Cui, Wenwen Tong, Kongzhi Hu, Jiapeng Luo, Zheng Ma, Ji Ma, Jiaqi Wang, Xiaoyi Dong, Hang Yan, Hewei Guo, Conghui He, Botian Shi, Zhenjiang Jin, Chao Xu, Bin Wang, Xingjian Wei, Wei Li, Wenjian Zhang, Bo Zhang, Pinlong Cai, Licheng Wen, Xiangchao Yan, Min Dou, Lewei Lu, Xizhou Zhu, Tong Lu, Dahua Lin, Yu Qiao, Jifeng Dai, and Wenhai Wang.
\newblock How far are we to gpt-4v? closing the gap to commercial multimodal models with open-source suites, 2024.

\bibitem[Deng et~al.(2025)Deng, Pan, Diao, Luo, Cui, Lu, Shan, Qi, and Wang]{deng2025autoregressive}
Haoge Deng, Ting Pan, Haiwen Diao, Zhengxiong Luo, Yufeng Cui, Huchuan Lu, Shiguang Shan, Yonggang Qi, and Xinlong Wang.
\newblock Autoregressive video generation without vector quantization.
\newblock In \emph{The Thirteenth International Conference on Learning Representations}, 2025.

\bibitem[{Gatis, Daniel} and contributors(2022)]{rembg}
{Gatis, Daniel} and contributors.
\newblock {rembg}: Image background removal tool, 2022.
\newblock GitHub repository.

\bibitem[{Google DeepMind}(2025)]{GoogleDeepMind2025Veo3}
{Google DeepMind}.
\newblock Veo: a text-to-video generation system.
\newblock Technical Report Veo 3 Tech Report, Google DeepMind, 2025.
\newblock Accessed: 2025-07-26.

\bibitem[HaCohen et~al.(2024)HaCohen, Chiprut, Brazowski, Shalem, Moshe, Richardson, Levin, Shiran, Zabari, Gordon, Panet, Weissbuch, Kulikov, Bitterman, Melumian, and Bibi]{HaCohen2024LTXVideo}
Yoav HaCohen, Nisan Chiprut, Benny Brazowski, Daniel Shalem, Dudu Moshe, Eitan Richardson, Eran Levin, Guy Shiran, Nir Zabari, Ori Gordon, Poriya Panet, Sapir Weissbuch, Victor Kulikov, Yaki Bitterman, Zeev Melumian, and Ofir Bibi.
\newblock Ltx-video: Realtime video latent diffusion, 2024.

\bibitem[He et~al.(2024)He, Jiang, Zhang, Ku, Soni, Siu, Chen, Chandra, Jiang, Arulraj, Wang, Do, Ni, Lyu, Narsupalli, Fan, Lyu, Lin, and Chen]{he-etal-2024-videoscore}
Xuan He, Dongfu Jiang, Ge Zhang, Max Ku, Achint Soni, Sherman Siu, Haonan Chen, Abhranil Chandra, Ziyan Jiang, Aaran Arulraj, Kai Wang, Quy~Duc Do, Yuansheng Ni, Bohan Lyu, Yaswanth Narsupalli, Rongqi Fan, Zhiheng Lyu, Bill~Yuchen Lin, and Wenhu Chen.
\newblock {V}ideo{S}core: Building automatic metrics to simulate fine-grained human feedback for video generation.
\newblock In \emph{Proceedings of the 2024 Conference on Empirical Methods in Natural Language Processing}, pages 2105--2123, Miami, Florida, USA, 2024. Association for Computational Linguistics.

\bibitem[Henschel et~al.(2025)Henschel, Khachatryan, Poghosyan, Hayrapetyan, Tadevosyan, Wang, Navasardyan, and Shi]{Henschel_2025_CVPR}
Roberto Henschel, Levon Khachatryan, Hayk Poghosyan, Daniil Hayrapetyan, Vahram Tadevosyan, Zhangyang Wang, Shant Navasardyan, and Humphrey Shi.
\newblock Streamingt2v: Consistent, dynamic, and extendable long video generation from text.
\newblock In \emph{Proceedings of the IEEE/CVF Conference on Computer Vision and Pattern Recognition (CVPR)}, pages 2568--2577, 2025.

\bibitem[Hessel et~al.(2021)Hessel, Holtzman, Forbes, Le~Bras, and Choi]{hessel-etal-2021-clipscore}
Jack Hessel, Ari Holtzman, Maxwell Forbes, Ronan Le~Bras, and Yejin Choi.
\newblock {CLIPS}core: A reference-free evaluation metric for image captioning.
\newblock In \emph{Proceedings of the 2021 Conference on Empirical Methods in Natural Language Processing}, pages 7514--7528, Online and Punta Cana, Dominican Republic, 2021. Association for Computational Linguistics.

\bibitem[Huang(2024)]{Huang2024VBenchSlides}
Ziqi Huang.
\newblock Comprehensive benchmark suite for video generative models: Vbench.
\newblock CVPR 2024 Workshop slides, 2024.
\newblock Accessed: 2025-11-12.

\bibitem[Huang et~al.(2024{\natexlab{a}})Huang, He, Yu, Zhang, Si, Jiang, Zhang, Wu, Jin, Chanpaisit, Wang, Chen, Wang, Lin, Qiao, and Liu]{huang2023vbench}
Ziqi Huang, Yinan He, Jiashuo Yu, Fan Zhang, Chenyang Si, Yuming Jiang, Yuanhan Zhang, Tianxing Wu, Qingyang Jin, Nattapol Chanpaisit, Yaohui Wang, Xinyuan Chen, Limin Wang, Dahua Lin, Yu Qiao, and Ziwei Liu.
\newblock {VBench}: Comprehensive benchmark suite for video generative models.
\newblock In \emph{Proceedings of the IEEE/CVF Conference on Computer Vision and Pattern Recognition}, 2024{\natexlab{a}}.

\bibitem[Huang et~al.(2024{\natexlab{b}})Huang, Zhang, Xu, He, Yu, Dong, Ma, Chanpaisit, Si, Jiang, Wang, Chen, Chen, Wang, Lin, Qiao, and Liu]{huang2024vbenchcomprehensiveversatilebenchmark}
Ziqi Huang, Fan Zhang, Xiaojie Xu, Yinan He, Jiashuo Yu, Ziyue Dong, Qianli Ma, Nattapol Chanpaisit, Chenyang Si, Yuming Jiang, Yaohui Wang, Xinyuan Chen, Ying-Cong Chen, Limin Wang, Dahua Lin, Yu Qiao, and Ziwei Liu.
\newblock Vbench++: Comprehensive and versatile benchmark suite for video generative models, 2024{\natexlab{b}}.

\bibitem[Jin et~al.(2025)Jin, Sun, Li, Xu, Xu, Jiang, Zhuang, Huang, Song, MU, and Lin]{jin2025pyramidal}
Yang Jin, Zhicheng Sun, Ningyuan Li, Kun Xu, Kun Xu, Hao Jiang, Nan Zhuang, Quzhe Huang, Yang Song, Yadong MU, and Zhouchen Lin.
\newblock Pyramidal flow matching for efficient video generative modeling.
\newblock In \emph{The Thirteenth International Conference on Learning Representations}, 2025.

\bibitem[Koukounas et~al.(2025)Koukounas, Mastrapas, Eslami, Wang, Akram, Günther, Mohr, Sturua, Wang, and Xiao]{DBLP:journals/corr/abs-2412-08802}
Andreas Koukounas, Georgios Mastrapas, Sedigheh Eslami, Bo Wang, Mohammad~Kalim Akram, Michael Günther, Isabelle Mohr, Saba Sturua, Nan Wang, and Han Xiao.
\newblock jina-clip-v2: Multilingual multimodal embeddings for text and images, 2025.

\bibitem[Krishna et~al.(2017)Krishna, Hata, Ren, Fei{-}Fei, and Niebles]{DBLP:conf/iccv/KrishnaHRFN17}
Ranjay Krishna, Kenji Hata, Frederic Ren, Li Fei{-}Fei, and Juan~Carlos Niebles.
\newblock Dense-captioning events in videos.
\newblock In \emph{{IEEE} International Conference on Computer Vision, {ICCV} 2017, Venice, Italy, October 22-29, 2017}, pages 706--715. {IEEE} Computer Society, 2017.

\bibitem[Kuaishou(2025)]{kling}
Kuaishou.
\newblock {Kling AI}, 2025.

\bibitem[Kwon et~al.(2023)Kwon, Li, Zhuang, Sheng, Zheng, Yu, Gonzalez, Zhang, and Stoica]{kwon2023efficient}
Woosuk Kwon, Zhuohan Li, Siyuan Zhuang, Ying Sheng, Lianmin Zheng, Cody~Hao Yu, Joseph~E. Gonzalez, Hao Zhang, and Ion Stoica.
\newblock Efficient memory management for large language model serving with pagedattention.
\newblock In \emph{Proceedings of the ACM SIGOPS 29th Symposium on Operating Systems Principles}, 2023.

\bibitem[Lee et~al.(2024)Lee, Kim, Park, Kim, and Seo]{lee-etal-2024-prometheus}
Seongyun Lee, Seungone Kim, Sue Park, Geewook Kim, and Minjoon Seo.
\newblock Prometheus-vision: Vision-language model as a judge for fine-grained evaluation.
\newblock In \emph{Findings of the Association for Computational Linguistics: ACL 2024}, pages 11286--11315, Bangkok, Thailand, 2024. Association for Computational Linguistics.

\bibitem[Liu and Zhang(2025)]{liu2025videolanguagemodelreliable}
Ming Liu and Wensheng Zhang.
\newblock Is your video language model a reliable judge?, 2025.

\bibitem[Liu et~al.(2023{\natexlab{a}})Liu, Zeng, Ren, Li, Zhang, Yang, Li, Yang, Su, Zhu, and Zhang]{liu2023grounding}
Shilong Liu, Zhaoyang Zeng, Tianhe Ren, Feng Li, Hao Zhang, Jie Yang, Chunyuan Li, Jianwei Yang, Hang Su, Jun Zhu, and Lei Zhang.
\newblock Grounding dino: Marrying dino with grounded pre-training for open-set object detection, 2023{\natexlab{a}}.

\bibitem[Liu et~al.(2025{\natexlab{a}})Liu, Zeng, Ren, Li, Zhang, Yang, Jiang, Li, Yang, Su, Zhu, and Zhang]{10.1007/978-3-031-72970-6_3}
Shilong Liu, Zhaoyang Zeng, Tianhe Ren, Feng Li, Hao Zhang, Jie Yang, Qing Jiang, Chunyuan Li, Jianwei Yang, Hang Su, Jun Zhu, and Lei Zhang.
\newblock Grounding dino: Marrying dino with grounded pre-training for open-set object detection.
\newblock In \emph{Computer Vision -- ECCV 2024}, pages 38--55, Cham, 2025{\natexlab{a}}. Springer Nature Switzerland.

\bibitem[Liu et~al.(2023{\natexlab{b}})Liu, Li, Ren, Gao, Li, Chen, Sun, and Hou]{liu2023fetv}
Yuanxin Liu, Lei Li, Shuhuai Ren, Rundong Gao, Shicheng Li, Sishuo Chen, Xu Sun, and Lu Hou.
\newblock {FETV}: A benchmark for fine-grained evaluation of open-domain text-to-video generation.
\newblock In \emph{Thirty-seventh Conference on Neural Information Processing Systems Datasets and Benchmarks Track}, 2023{\natexlab{b}}.

\bibitem[Liu et~al.(2024)Liu, Cun, Liu, Wang, Zhang, Chen, Liu, Zeng, Chan, and Shan]{Liu_2024_CVPR}
Yaofang Liu, Xiaodong Cun, Xuebo Liu, Xintao Wang, Yong Zhang, Haoxin Chen, Yang Liu, Tieyong Zeng, Raymond Chan, and Ying Shan.
\newblock Evalcrafter: Benchmarking and evaluating large video generation models.
\newblock In \emph{Proceedings of the IEEE/CVF Conference on Computer Vision and Pattern Recognition (CVPR)}, pages 22139--22149, 2024.

\bibitem[Liu et~al.(2025{\natexlab{b}})Liu, Zhu, Ren, Wang, Guo, Sun, and Jiang]{liu2025uvemllmsunifiedevaluators}
Yuanxin Liu, Rui Zhu, Shuhuai Ren, Jiacong Wang, Haoyuan Guo, Xu Sun, and Lu Jiang.
\newblock Uve: Are mllms unified evaluators for ai-generated videos?, 2025{\natexlab{b}}.

\bibitem[{mertcobanov} and contributors(2024)]{nature_dataset_hf}
{mertcobanov} and contributors.
\newblock {Nature Dataset}: Landscape background images, 2024.
\newblock Hugging Face dataset; ID: mertcobanov/nature-dataset.

\bibitem[Nagrani et~al.(2025)Nagrani, Zhang, Mehran, Hornung, Gundavarapu, Jha, Myers, Zhou, Gong, Schmid, Sirotenko, Zhu, and Weyand]{nagrani2025neptunelongorbitbenchmarking}
Arsha Nagrani, Mingda Zhang, Ramin Mehran, Rachel Hornung, Nitesh~Bharadwaj Gundavarapu, Nilpa Jha, Austin Myers, Xingyi Zhou, Boqing Gong, Cordelia Schmid, Mikhail Sirotenko, Yukun Zhu, and Tobias Weyand.
\newblock Neptune: The long orbit to benchmarking long video understanding, 2025.

\bibitem[OpenAI(2025)]{OpenAI2025GPT5SystemCard}
OpenAI.
\newblock Gpt-5 system card.
\newblock Technical report, OpenAI, 2025.
\newblock Accessed: 2025-11-12.

\bibitem[Qian et~al.(2024)]{qian2024streaming}
Rui Qian et~al.
\newblock Streaming long video understanding with large language models.
\newblock In \emph{Advances in Neural Information Processing Systems}, 2024.

\bibitem[Qin et~al.(2020)Qin, Zhang, Huang, Dehghan, Zaiane, and Jagersand]{QIN2020107404}
Xuebin Qin, Zichen Zhang, Chenyang Huang, Masood Dehghan, Osmar~R. Zaiane, and Martin Jagersand.
\newblock U2-net: Going deeper with nested u-structure for salient object detection.
\newblock \emph{Pattern Recognition}, 106:\penalty0 107404, 2020.

\bibitem[Rombach et~al.(2022)Rombach, Blattmann, Lorenz, Esser, and Ommer]{Rombach_2022_CVPR}
Robin Rombach, Andreas Blattmann, Dominik Lorenz, Patrick Esser, and Bj\"orn Ommer.
\newblock High-resolution image synthesis with latent diffusion models.
\newblock In \emph{Proceedings of the IEEE/CVF Conference on Computer Vision and Pattern Recognition (CVPR)}, pages 10684--10695, 2022.

\bibitem[Singer et~al.(2023)Singer, Polyak, Hayes, Yin, An, Zhang, Hu, Yang, Ashual, Gafni, Parikh, Gupta, and Taigman]{singer2023makeavideo}
Uriel Singer, Adam Polyak, Thomas Hayes, Xi Yin, Jie An, Songyang Zhang, Qiyuan Hu, Harry Yang, Oron Ashual, Oran Gafni, Devi Parikh, Sonal Gupta, and Yaniv Taigman.
\newblock Make-a-video: Text-to-video generation without text-video data.
\newblock In \emph{The Eleventh International Conference on Learning Representations}, 2023.

\bibitem[Somers(2005)]{somers-2005-round}
Harold Somers.
\newblock Round-trip translation: What is it good for?
\newblock In \emph{Proceedings of the Australasian Language Technology Workshop 2005}, pages 127--133, Sydney, Australia, 2005.

\bibitem[Sun et~al.(2025)Sun, Huang, Liu, Wu, Xu, Li, and Liu]{Sun_2025_CVPR}
Kaiyue Sun, Kaiyi Huang, Xian Liu, Yue Wu, Zihan Xu, Zhenguo Li, and Xihui Liu.
\newblock T2v-compbench: A comprehensive benchmark for compositional text-to-video generation.
\newblock In \emph{Proceedings of the IEEE/CVF Conference on Computer Vision and Pattern Recognition (CVPR)}, pages 8406--8416, 2025.

\bibitem[Tapaswi et~al.(2016)Tapaswi, Zhu, Stiefelhagen, Torralba, Urtasun, and Fidler]{Tapaswi_2016_CVPR}
Makarand Tapaswi, Yukun Zhu, Rainer Stiefelhagen, Antonio Torralba, Raquel Urtasun, and Sanja Fidler.
\newblock Movieqa: Understanding stories in movies through question-answering.
\newblock In \emph{Proceedings of the IEEE Conference on Computer Vision and Pattern Recognition (CVPR)}, 2016.

\bibitem[Team(2025)]{qwen3technicalreport}
Qwen Team.
\newblock Qwen3 technical report, 2025.

\bibitem[Tian et~al.(2025)Tian, Zou, Yang, and Zhang]{Tian_2025_CVPR}
Xinyu Tian, Shu Zou, Zhaoyuan Yang, and Jing Zhang.
\newblock Identifying and mitigating position bias of multi-image vision-language models.
\newblock In \emph{Proceedings of the IEEE/CVF Conference on Computer Vision and Pattern Recognition (CVPR)}, pages 10599--10609, 2025.

\bibitem[Tomar(2006)]{tomar2006converting}
Suramya Tomar.
\newblock Converting video formats with ffmpeg.
\newblock \emph{Linux Journal}, 2006\penalty0 (146):\penalty0 10, 2006.

\bibitem[Villegas et~al.(2023)Villegas, Babaeizadeh, Kindermans, Moraldo, Zhang, Saffar, Castro, Kunze, and Erhan]{villegas2023phenaki}
Ruben Villegas, Mohammad Babaeizadeh, Pieter-Jan Kindermans, Hernan Moraldo, Han Zhang, Mohammad~Taghi Saffar, Santiago Castro, Julius Kunze, and Dumitru Erhan.
\newblock Phenaki: Variable length video generation from open domain textual descriptions.
\newblock In \emph{International Conference on Learning Representations}, 2023.

\bibitem[Wang et~al.(2024{\natexlab{a}})Wang, Bai, Tan, Wang, Fan, Bai, Chen, Liu, Wang, Ge, Fan, Dang, Du, Ren, Men, Liu, Zhou, Zhou, and Lin]{Qwen2VL}
Peng Wang, Shuai Bai, Sinan Tan, Shijie Wang, Zhihao Fan, Jinze Bai, Keqin Chen, Xuejing Liu, Jialin Wang, Wenbin Ge, Yang Fan, Kai Dang, Mengfei Du, Xuancheng Ren, Rui Men, Dayiheng Liu, Chang Zhou, Jingren Zhou, and Junyang Lin.
\newblock Qwen2-vl: Enhancing vision-language model's perception of the world at any resolution.
\newblock \emph{arXiv preprint arXiv:2409.12191}, 2024{\natexlab{a}}.

\bibitem[Wang et~al.(2024{\natexlab{b}})Wang, He, Hong, Cheng, Zhang, Qi, Gu, Huang, Xu, Dong, Ding, and Tang]{wang2024lvbenchextremelongvideo}
Weihan Wang, Zehai He, Wenyi Hong, Yean Cheng, Xiaohan Zhang, Ji Qi, Xiaotao Gu, Shiyu Huang, Bin Xu, Yuxiao Dong, Ming Ding, and Jie Tang.
\newblock Lvbench: An extreme long video understanding benchmark, 2024{\natexlab{b}}.

\bibitem[Wu et~al.(2025)Wu, Li, Zhou, Lin, Gao, Yan, ming Yin, Bai, Xu, Chen, Chen, Tang, Zhang, Wang, Yang, Yu, Cheng, Liu, Li, Zhang, Meng, Wei, Ni, Chen, Cao, Peng, Qu, Wu, Wang, Yu, Wen, Feng, Xu, Wang, Zhang, Zhu, Wu, Cai, and Liu]{wu2025qwenimagetechnicalreport}
Chenfei Wu, Jiahao Li, Jingren Zhou, Junyang Lin, Kaiyuan Gao, Kun Yan, Sheng ming Yin, Shuai Bai, Xiao Xu, Yilei Chen, Yuxiang Chen, Zecheng Tang, Zekai Zhang, Zhengyi Wang, An Yang, Bowen Yu, Chen Cheng, Dayiheng Liu, Deqing Li, Hang Zhang, Hao Meng, Hu Wei, Jingyuan Ni, Kai Chen, Kuan Cao, Liang Peng, Lin Qu, Minggang Wu, Peng Wang, Shuting Yu, Tingkun Wen, Wensen Feng, Xiaoxiao Xu, Yi Wang, Yichang Zhang, Yongqiang Zhu, Yujia Wu, Yuxuan Cai, and Zenan Liu.
\newblock Qwen-image technical report, 2025.

\bibitem[Wu et~al.(2024)Wu, Li, Chen, and Li]{NEURIPS2024_329ad516}
Haoning Wu, Dongxu Li, Bei Chen, and Junnan Li.
\newblock Longvideobench: A benchmark for long-context interleaved video-language understanding.
\newblock In \emph{Advances in Neural Information Processing Systems}, pages 28828--28857. Curran Associates, Inc., 2024.

\bibitem[Yang et~al.(2025)Yang, Yu, Li, Liu, Huang, Huang, Jiang, Tu, Zhang, Zhou, Lin, Dang, Yang, Yu, Li, Sun, Zhu, Men, He, Xu, Yin, Yu, Qiu, Ren, Yang, Li, Xu, and Zhang]{qwen2.5-1m}
An Yang, Bowen Yu, Chengyuan Li, Dayiheng Liu, Fei Huang, Haoyan Huang, Jiandong Jiang, Jianhong Tu, Jianwei Zhang, Jingren Zhou, Junyang Lin, Kai Dang, Kexin Yang, Le Yu, Mei Li, Minmin Sun, Qin Zhu, Rui Men, Tao He, Weijia Xu, Wenbiao Yin, Wenyuan Yu, Xiafei Qiu, Xingzhang Ren, Xinlong Yang, Yong Li, Zhiying Xu, and Zipeng Zhang.
\newblock Qwen2.5-1m technical report.
\newblock \emph{arXiv preprint arXiv:2501.15383}, 2025.

\bibitem[Yang et~al.(2024)Yang, Huang, Lu, Han, Zhang, Gao, Hu, and Zhao]{NEURIPS2024_6903a5aa}
Dongjie Yang, Suyuan Huang, Chengqiang Lu, Xiaodong Han, Haoxin Zhang, Yan Gao, Yao Hu, and Hai Zhao.
\newblock Vript: A video is worth thousands of words.
\newblock In \emph{Advances in Neural Information Processing Systems}, pages 57240--57261. Curran Associates, Inc., 2024.

\bibitem[Zeng et~al.(2025)Zeng, Li, Wang, Li, Jiang, Yan, Li, Shi, Yue, Wang, Wang, Qiao, and Wang]{zeng2025timesuite}
Xiangyu Zeng, Kunchang Li, Chenting Wang, Xinhao Li, Tianxiang Jiang, Ziang Yan, Songze Li, Yansong Shi, Zhengrong Yue, Yi Wang, Yali Wang, Yu Qiao, and Limin Wang.
\newblock Timesuite: Improving mllms for long video understanding via grounded tuning.
\newblock In \emph{International Conference on Learning Representations}, 2025.

\bibitem[Zhang et~al.(2023)Zhang, Li, and Bing]{zhang-etal-2023-video}
Hang Zhang, Xin Li, and Lidong Bing.
\newblock Video-{LL}a{MA}: An instruction-tuned audio-visual language model for video understanding.
\newblock In \emph{Proceedings of the 2023 Conference on Empirical Methods in Natural Language Processing: System Demonstrations}, pages 543--553, Singapore, 2023. Association for Computational Linguistics.

\bibitem[Zheng et~al.(2025)Zheng, Huang, Liu, Zou, He, Zhang, Zhang, He, Zheng, Qiao, and Liu]{zheng2025vbench20advancingvideogeneration}
Dian Zheng, Ziqi Huang, Hongbo Liu, Kai Zou, Yinan He, Fan Zhang, Yuanhan Zhang, Jingwen He, Wei-Shi Zheng, Yu Qiao, and Ziwei Liu.
\newblock Vbench-2.0: Advancing video generation benchmark suite for intrinsic faithfulness, 2025.

\bibitem[Zhou et~al.(2025)Zhou, Shu, Zhao, Wu, Liang, Xiao, Qin, Yang, Xiong, Zhang, Huang, and Liu]{Zhou_2025_CVPR}
Junjie Zhou, Yan Shu, Bo Zhao, Boya Wu, Zhengyang Liang, Shitao Xiao, Minghao Qin, Xi Yang, Yongping Xiong, Bo Zhang, Tiejun Huang, and Zheng Liu.
\newblock Mlvu: Benchmarking multi-task long video understanding.
\newblock In \emph{Proceedings of the IEEE/CVF Conference on Computer Vision and Pattern Recognition (CVPR)}, pages 13691--13701, 2025.

\end{thebibliography}
}

\clearpage
\setcounter{page}{1}
\maketitlesupplementary
\appendix
\newcommand{\lbl}[1]{\textsc{#1}}

\section{Human annotation and quality control}
\label{sec:appendix_annotation}
This section provides details of the crowdsourced annotation described in Section~\ref{subsec:human_annotation}.

\subsection{Annotation platform and crowd workers}

\paragraph{Platform.}
We collected all human annotations on Yahoo!~Crowdsourcing, a commercial crowdsourcing platform operated by Yahoo Japan Corporation.

\paragraph{Crowd workers.}
In total, 736 unique crowd workers completed 3{,}793 annotation tasks.
We recruited only general users registered on the platform and did not require any specific expert background.

\paragraph{Compensation.}
For each task, defined as evaluating a single video pair, we set the reward based on the expected time needed to watch and compare the two videos.
The unit price was set so that the effective hourly wage was higher than the legal average minimum wage in Japan.
Across the entire annotation project, the total payment was 408{,}520~JPY (approximately 2{,}685~USD at the time of the experiments).

\subsection{Collection of human accuracy scores}
As described in Section~\ref{subsec:meta_evaluation_criteria}, we use accuracy as the meta evaluation metric.
To obtain the human performance, we asked crowd workers to solve the same pairwise comparison task as the automatic evaluation systems.

Given a prompt $p$ and a video pair $\{v_{p}^{+}, v_{p}^{-}\}$, the interface showed the prompt and two video players side by side.
Since the original prompts were written in English while our crowd workers were Japanese, we translated each prompt into Japanese using Qwen3-Next-80B-A3B-Instruct-FP8~\cite{qwen3technicalreport,qwen2.5-1m} and presented the both original and translated prompt $p$.
We instructed workers to read the prompt $p$, watch both videos, and indicate which video they considered to be higher quality with respect to the target aspect.
The video player allowed workers to move the playhead using a seek bar and to replay arbitrary segments of each video, so that they could inspect any part of the content if necessary.

We computed human accuracy in the same way as for the evaluation systems, following the definition in Section~\ref{subsec:meta_evaluation_criteria}, that is, as the proportion of pairs in which the worker selected $v_{p}^{+}$.

\begin{table*}[t]
  \centering
  \small
  \caption{
    Filtering statistics for the annotated degraded pairs.
    For each aspect, we report the number of candidate samples before filtering (Initial), the number of samples excluded because at least one worker selected option C (Excluded ($C \neq 0$)), the number of samples excluded because the number of votes for option A did not exceed that for option B (Excluded ($A \leq B$)), the number of samples retained after filtering (Final), and the resulting retention rate (Retention Rate).
    The filtering conditions follow the human annotation procedure defined in Section~\ref{subsec:human_annotation}.
  }
  \label{tab:appendix_filtering_stats}
  \begin{tabular}{lrrrrr}
    \toprule
    \textbf{Aspect} & \textbf{Initial Samples} & \textbf{Excluded ($C \neq 0$)} & \textbf{Excluded ($A \leq B$)} & \textbf{Final Samples} & \textbf{Retention Rate (\%)} \\
    \midrule
    Aesthetics             & 446 &  92 &  72 & 282 & 63.3 \\
    Technical Quality      & 446 & 243 &  72 & 131 & 29.4 \\
    Appearance Style       & 446 & 112 &  22 & 312 & 70.0 \\
    Background Consistency & 446 &  75 &  17 & 354 & 79.4 \\
    Temporal Flow          & 444 & 115 &  44 & 285 & 64.2 \\
    Comprehensiveness      & 430 &  83 & 112 & 235 & 54.7 \\
    Object Integrity       & 375 &  96 & 179 & 100 & 26.7 \\
    Spatial Relationship   & 435 & 120 &  79 & 236 & 54.3 \\
    Dynamics Degree        & 444 &  89 &  22 & 333 & 75.0 \\
    Color                  & 327 &  56 &  67 & 204 & 62.4 \\
    \midrule
    \textbf{Total}         & 4239 & 1081 & 686 & 2472 & 58.3 \\
    \bottomrule
  \end{tabular}
\end{table*}

\subsection{Annotation UI for degradation checking}

\begin{figure}[t]
  \centering
  \adjincludegraphics[width=\linewidth,clip,trim={{0.55\textwidth} 0 {0.55\textwidth} {0}}]{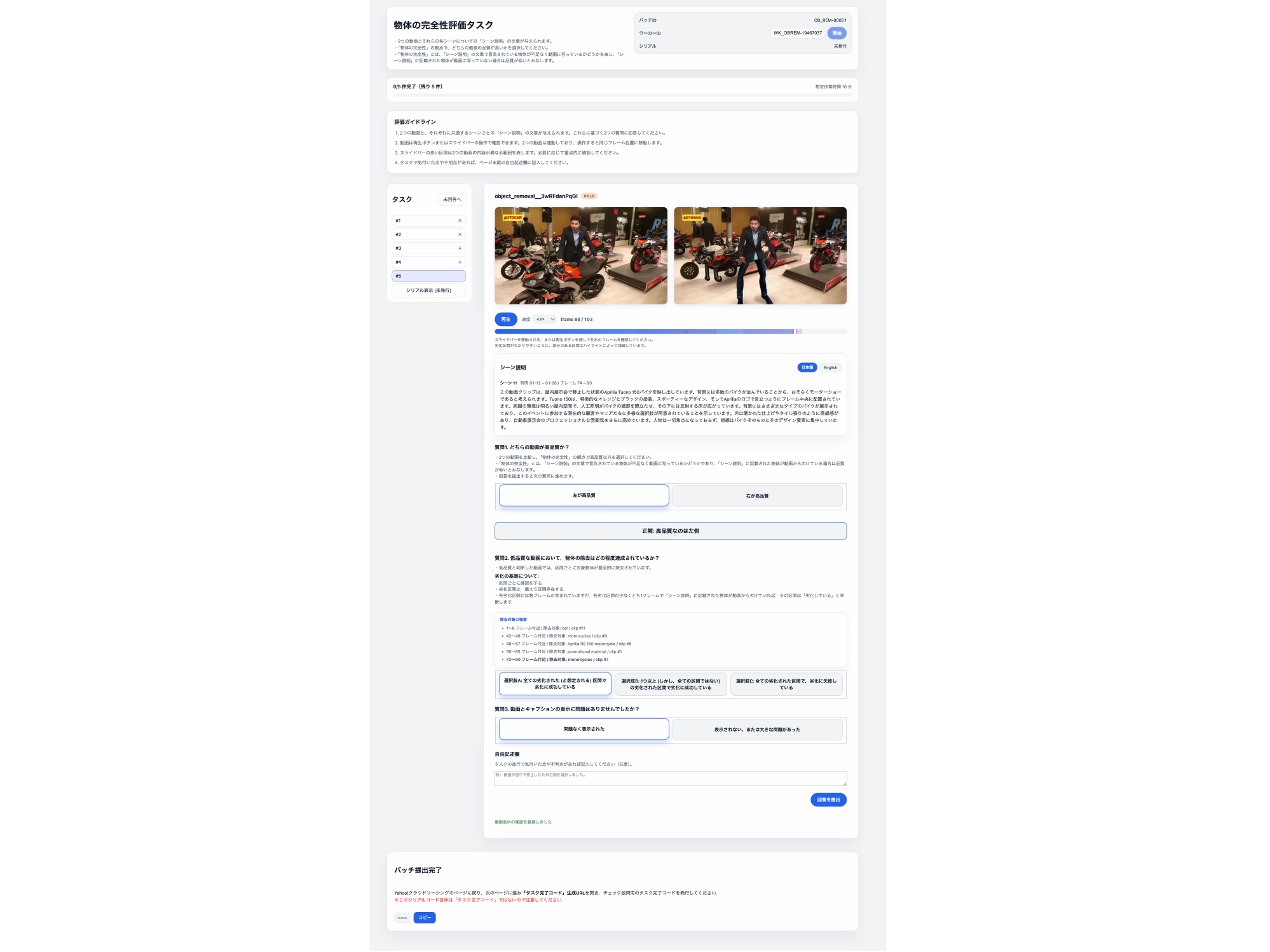}
  \caption{
    Example of the annotation interface used for the \textit{Object Integrity} aspect.
    The interface presents the prompt and two videos, collects the worker's choice of the higher quality video, and then asks additional questions to verify whether the synthetic degradation for the target aspect has been applied correctly.
  }
  \label{fig:annotation_ui}
\end{figure}

Figure~\ref{fig:annotation_ui} shows the annotation interface used for degradation checking.
In addition to the pairwise quality judgment, we collected A/B/C ratings that indicate whether the synthetic degradation succeeded for each aspect, as described in Section~\ref{subsec:human_annotation}.
After each crowd worker finishes judging which of the two videos is better, we present our intended gold label.
They then rate on a three-point A/B/C scale whether this gold label is appropriate.
For each video pair and aspect, five crowd workers provided A/B/C judgments with the following options:

\begin{itemize}[leftmargin=1.5em]
    \item[\textbf{A.}] The degradation clearly succeeds in all selected clips.
    \item[\textbf{B.}] The degradation succeeds in at least one but not all selected clips, or the effect is weak.
    \item[\textbf{C.}] The degradation fails in all selected clips.
\end{itemize}

\begin{table*}[t]
    \caption{
        Evaluation criteria and instructions provided to crowd workers for each aspect.
        The Japanese instructions were presented to the workers, and the English instructions are their translations.
    }
    \centering
    \small
    \begin{tabular}{p{0.16\linewidth} p{0.38\linewidth} p{0.38\linewidth}}
        \toprule
        \textbf{Aspect} & \textbf{Japanese Instruction} & \textbf{English Instruction} \\
        \midrule
        \textbf{Aesthetics} &
        「美的品質」とは，各フレームが構図・調和・写真的品質を含む総合的な視覚的魅力を有しているかどうかを表し，「シーン説明」との色合いに矛盾があったり，動画の見た目の好ましさが劣っている場合は品質が低いとみなします． &
        ``Aesthetics'' indicates whether each frame possesses overall visual appeal, including composition, harmony, and photographic quality. Quality is considered low if there is a tonal contradiction with the ``scene description'' or if the visual appeal of the video is inferior. \\
        \midrule
        \textbf{Technical Quality} &
        「技術的品質」とは，技術的な映像のミス（例：ぼけ・ブレ，ノイズ，低画質，輪郭歪み）がないかどうかを表し，技術的な映像ミスが存在する場合は品質が低いとみなします． &
        ``Technical Quality'' indicates whether the video is free from technical artifacts (e.g., blur/shake, noise, low resolution, contour distortion). Quality is considered low if such technical video errors exist. \\
        \midrule
        \textbf{Appearance Style} &
        「スタイル」とは，「シーン説明」で指定された外観スタイル（例：油絵風，漫画風，水彩画風，スケッチ風など）が動画内で適切に再現されているかどうかを表し，指定されたスタイルと異なる表現にとなっている場合は品質が低いとみなします． &
        ``Appearance Style'' indicates whether the visual style specified in the ``scene description'' (e.g., oil painting, cartoon, watercolor, sketch) is appropriately reproduced in the video. Quality is considered low if the representation differs from the specified style. \\
        \midrule
        \textbf{Background \newline{Consistency}} &
        「背景の一貫性」とは，フレーム間で背景が視覚的に安定かつ一貫して保たれているかどうかを表し，連続フレームで背景が一貫していない動画は品質が低いとみなします． &
        ``Background Consistency'' indicates whether the background remains visually stable and consistent across frames. Quality is considered low if the background is inconsistent across consecutive frames. \\
        \midrule
        \textbf{Object Integrity} &
        「物体の完全性」とは，「シーン説明」の文章で言及されている物体が不足なく動画に写っているかどうかを表し，「シーン説明」に記載された物体が動画に写っていない場合は品質が低いとみなします． &
        ``Object Integrity'' indicates whether the objects mentioned in the ``scene description'' are present in the video without omission. Quality is considered low if objects described in the ``scene description'' do not appear in the video. \\
        \midrule
        \textbf{Color} &
        「色」とは，各フレーム内における色が「シーン説明」で指定された色（例：物体の色など）と一致しているかどうかを表し，指定された色と異なる場合は品質が低いとみなします． &
        ``Color'' indicates whether the colors within frames match those specified in the ``scene description'' (e.g., object colors). Quality is considered low if the colors differ from the specifications. \\
        \midrule
        \textbf{Dynamics Degree} &
        「動きの度合い」とは，各フレームに含まれる動きの量が適切かどうかを表し，「シーン説明」の内容に比較して静的すぎたり，過度に不規則な場合は品質が低いとみなします． &
        ``Dynamics Degree'' indicates whether the amount of motion in each frame is appropriate. Quality is considered low if the video is too static or excessively erratic compared to the content of the ``scene description.'' \\
        \midrule
        \textbf{Comprehensiveness} &
        「網羅性」とは，「シーン説明」の内容が動画に反映されているかどうかを表し，「シーン説明」に記述されているシーンが動画内で適切に表示されない場合には品質が低いとみなします． &
        ``Comprehensiveness'' indicates whether the content of the ``scene description'' is reflected in the video. Quality is considered low if scenes described in the ``scene description'' are not appropriately displayed within the video. \\
        \midrule
        \textbf{Spatial \newline{Relationship}} &
        「空間的関係」とは，各フレームの物体が「シーン説明」で示された空間的関係（左右/上下/前後/近接）と一致しているかどうかを表し，いづれかの空間関係に矛盾が生じている場合は品質が低いとみなします． &
        ``Spatial Relationship'' indicates whether the objects in each frame match the spatial relationships (left/right, up/down, front/back, proximity) indicated in the ``scene description.'' Quality is considered low if a contradiction arises in any spatial relationship. \\
        \midrule
        \textbf{Temporal Flow} &
        「時系列の流れ」とは，「シーン説明」の文章のイベントの発生順番と動画内のシーンの順番が適切に対応しているかを表し，「シーン説明」の内容に矛盾した無関係なイベントが発生する動画は品質が低いとみなします． &
        ``Temporal Flow'' indicates whether the order of events in the ``scene description'' text corresponds appropriately to the order of scenes in the video. Quality is considered low if irrelevant events occur or if the order contradicts the ``scene description.'' \\
        \bottomrule
    \end{tabular}
    \label{tab:crowd_worker_instruction}
\end{table*}

Table~\ref{tab:crowd_worker_instruction} lists the evaluation criteria shown to crowd workers for each aspect of the human evaluation.

\paragraph{Viewing enforcement.}
Long-video evaluation tasks often yield unreliable answers when workers skip large portions of the videos.
To reduce this risk, we introduced a viewing enforcement mechanism in the degradation-checking UI.
For each pair, we predefined the clip interval where the degradation was applied.
In the interface, workers could freely move the seek bar, but the A/B/C response buttons for degradation success remained disabled until the playhead had been moved to the designated degraded interval and the worker clicked a confirmation button indicating that they had checked the degraded region.
This design made skipping behavior difficult.

\subsection{Crowd worker filtering}
We additionally applied worker-level filtering to reduce the impact of unreliable crowd workers.
We monitored response patterns across tasks and identified workers who repeatedly exhibited low-effort behavior.
We used the following indicators.

\begin{enumerate}
    \item \textbf{Abnormally short completion times.}
    We flagged workers whose task completion times were extremely short relative to the total duration of the video pair, for example those consistently in the fastest 10\% for a given task configuration.
    \item \textbf{Persistent disagreement on easy cases.}
    For samples where the agreement among other workers was very high (for example, 4 out of 5 workers chose A), we flagged workers who repeatedly gave clearly inconsistent answers (for example, answering C) on such samples.
    \item \textbf{Very low task level accuracy.}
    For tasks where the majority of workers achieved high accuracy (around 80\% or higher), we flagged workers whose accuracy remained very low (20\% or lower) over many assignments.
\end{enumerate}

Workers who exhibited problematic response patterns according to multiple indicators were manually reviewed.
Workers judged to be unreliable were excluded from later task assignments.
By the time we finished collecting annotations for the last aspect, this filtering process had excluded 227 workers in total from annotation tasks.

\begin{figure*}[t]
  \centering
  \adjincludegraphics[
    width=0.96\textwidth,
    clip,
    trim={{.0\width} {.33\height} {.0\width} {0}}
  ]{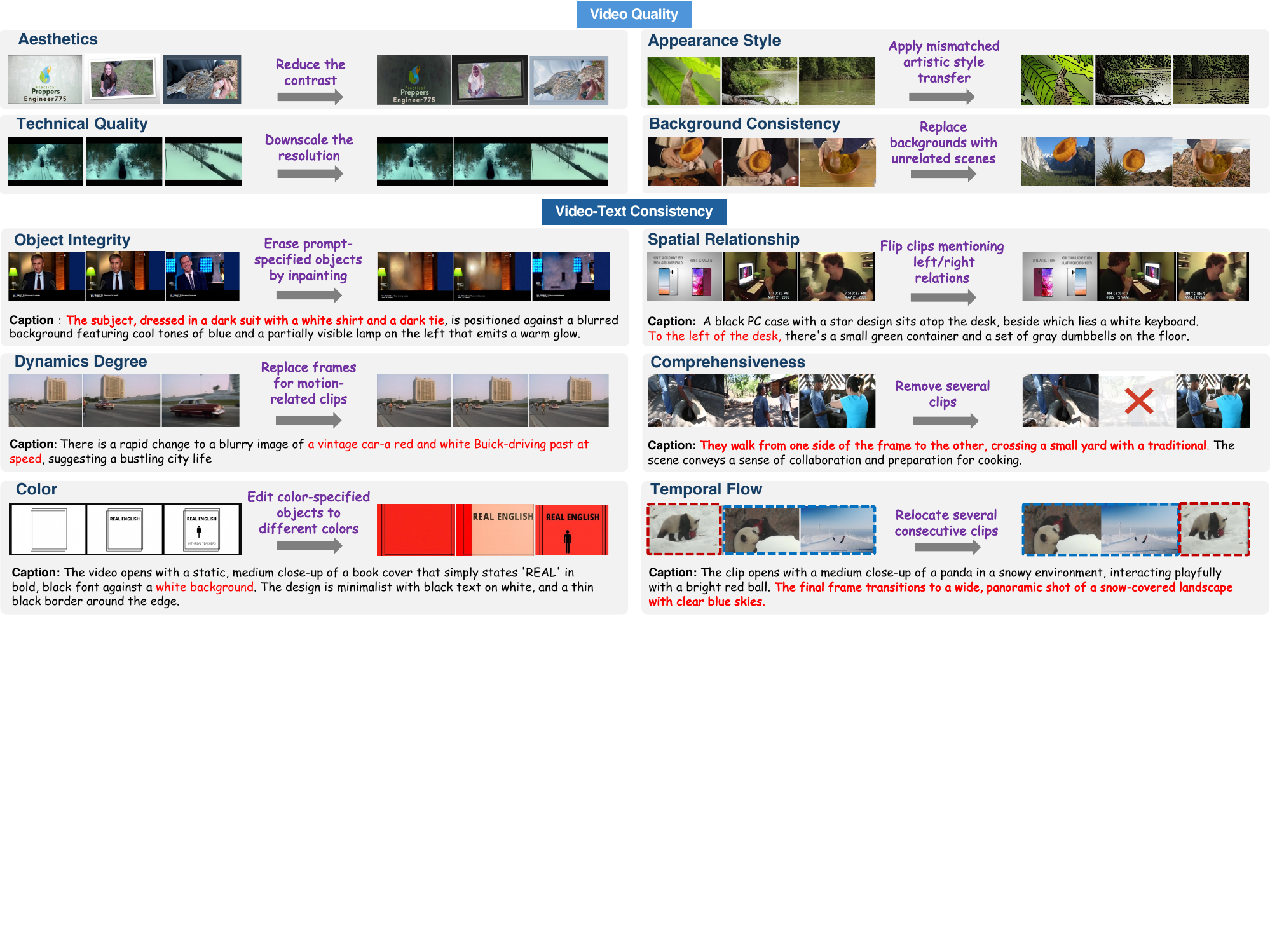}
  \caption{
    Representative examples of original and degraded video pairs in SLVMEval.
    For each of the 10 aspects, we show one original video (left) and its aspect-specifically degraded video (right).
    The upper rows correspond to the \textit{Video Quality} aspects, and the lower rows correspond to the \textit{Video-Text Consistency} aspects.
    Frames are sampled around the clips where the synthetic degradation is applied.
  }
  \label{fig:10_example_videos}
\end{figure*}

\subsection{Filtering statistics}
We applied the filtering criteria defined in Section~\ref{subsec:human_annotation} to the A/B/C degradation ratings to construct the final testbed.
Concretely, we retained a sample only if it satisfied both
\[
\text{count}(C) = 0 \quad \text{and} \quad \text{count}(A) > \text{count}(B),
\]
where $\text{count}(A)$, $\text{count}(B)$, and $\text{count}(C)$ denote the numbers of workers who chose A, B, and C, respectively.

Table~\ref{tab:appendix_filtering_stats} summarizes the statistics of this filtering process for each aspect.
For each aspect, the table reports the number of candidate samples before filtering (Initial), the number of samples excluded because at least one worker selected option C (Excluded ($C \neq 0$)), the number of samples excluded because the number of votes for A was not greater than the number of votes for B (Excluded ($A \leq B$)), the number of samples retained after filtering (Final), and the resulting retention rate.
Under these conditions, at least 100 long video pairs remain for every aspect.

\section{Examples of degraded video pairs}
Figure~\ref{fig:10_example_videos} shows example pairs of original and degraded videos, one for each of the ten aspects in SLVMEval.
For each aspect, we show one source long video and its degraded counterpart, constructed using the aspect-specific operation described in Section~\ref{subsec:degradation_operations}.
In each pair, the left column displays frames from the original video and the right column displays frames from the degraded video.
We extract frames from around the clips to which the degradation is applied.

The aspects in the \textit{Video Quality} category (Aesthetics, Technical Quality, Appearance Style, Background Consistency) appear in the top row, and those in the \textit{Video-Text Consistency} category (Object Integrity, Color, Dynamics Degree, Comprehensiveness, Spatial Relationship, Temporal Flow) appear in the bottom row.

\section{Proxy validation on generated videos}
\label{appendix:sec:additional_experiment}

\begin{figure}[t]
    \centering
    \includegraphics[width=0.90\linewidth]{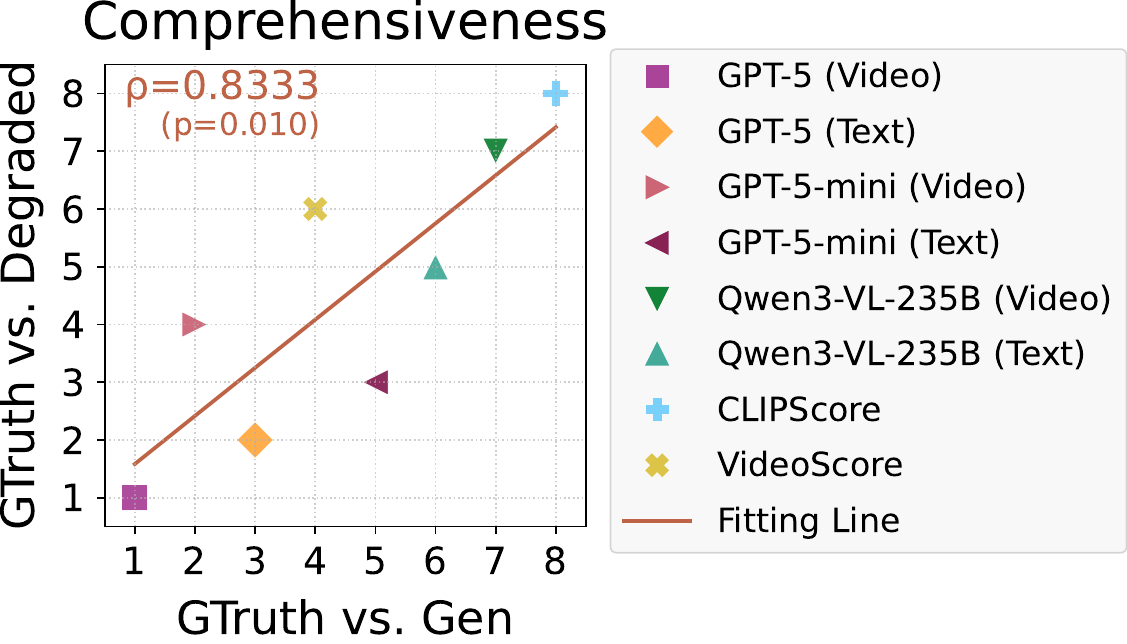}
    \caption{Rank consistency on FETV.}
    \label{fig:human_vs_degrade_rank}
\end{figure}

Directly validating SLVMEval on real long generated videos is currently difficult, as existing text-to-long video generation systems do not yet reliably produce videos at the target scale by SLVMEval with sufficiently stable quality.
Therefore, we conduct a proxy experiment on short videos to examine whether the relative ranking of evaluation systems induced by our synthetic degradations is consistent with the ranking obtained in a generated-video comparison setting.

\subsection{Dataset and settings}
We use a subset of FETV~\cite{liu2023fetv}, which contains short videos lasting a few seconds with human annotations for multiple aspects, including \emph{alignment}.
FETV provides pairs of human-created videos (\lbl{GTruth}) and generated videos (\lbl{Gen}), together with human quality signals.
We also construct \lbl{Degraded} videos by applying our degradation operations to \lbl{GTruth}.
Since our degradation operations assume a multi-clip structure (Algorithm~\ref{alg:phi_low_overview}), we concatenate two videos and treat each as a clip.

We consider two pairwise meta-evaluation settings:
(i) \lbl{GTruth} vs.\ \lbl{Gen}; the video with a higher human score is regarded as higher quality,
(ii) \lbl{GTruth} vs.\ \lbl{Degraded}; the ground truth is always regarded as higher quality.
For each setting, we evaluate each system using pairwise accuracy (\S~3.2), rank systems by accuracy, and compute the Spearman rank correlation between the two system rankings.

\subsection{Results and interpretation}
We focus on the FETV aspect \emph{alignment}, which most closely corresponds to our \textbf{Comprehensiveness} aspect.
As shown in Figure~\ref{fig:human_vs_degrade_rank}, the system ranking derived from \lbl{GTruth} vs.\ \lbl{Degraded} is strongly correlated with the ranking derived from \lbl{GTruth} vs.\ \lbl{Gen} ($\rho=0.8333$).
Although limited to a single aspect and short videos, this suggests that the properties of the generated videos and our degraded videos are unlikely to be overly dissimilar.

\section{Dataset statistics}
\label{subsec:dataset_statistics}

\begin{figure}[t]
  \centering
  \includegraphics[width=0.9\linewidth]{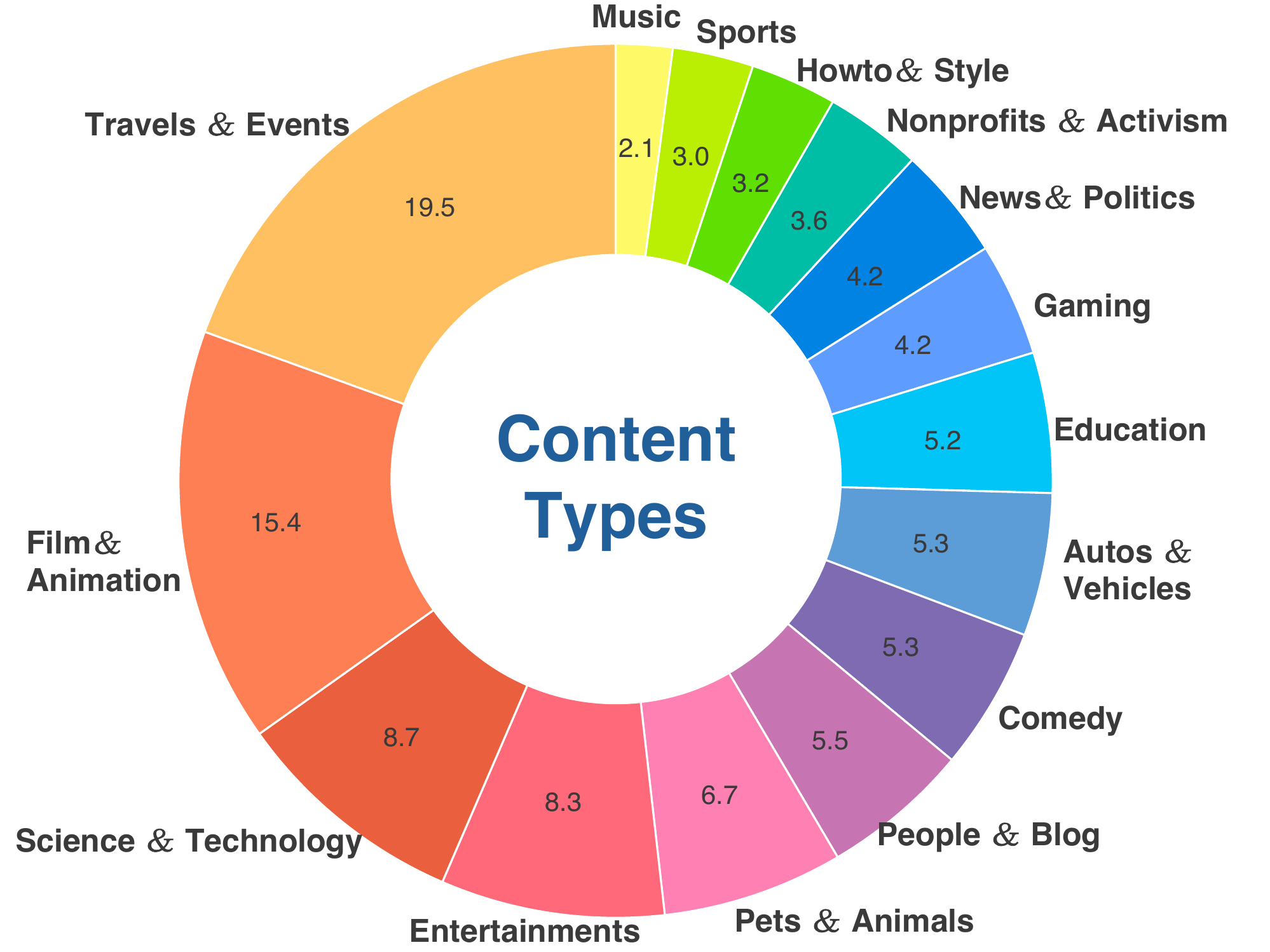}
  \caption{Distribution of video content categories in SLVMEval, showing a balanced representation across 15 categories.}
  \label{fig:video_genre}
\end{figure}

Figure~\ref{fig:video_genre} summarizes the distribution of video content categories in SLVMEval.
As described in Section~\ref{subsec:dataset_source}, we construct the benchmark from a dense video captioning dataset that spans diverse real-world topics, and then apply aspect-specific degrading operations followed by human filtering.
The resulting test set covers 15 distinct categories, including everyday activities, nature, sports, indoor scenes, and others.
No single category dominates the dataset, and the categories are approximately balanced so that the benchmark evaluates robustness to long-video evaluation without bias toward a particular type of content.

\begin{figure}[t]
  \centering
  \includegraphics[width=1.0\linewidth]{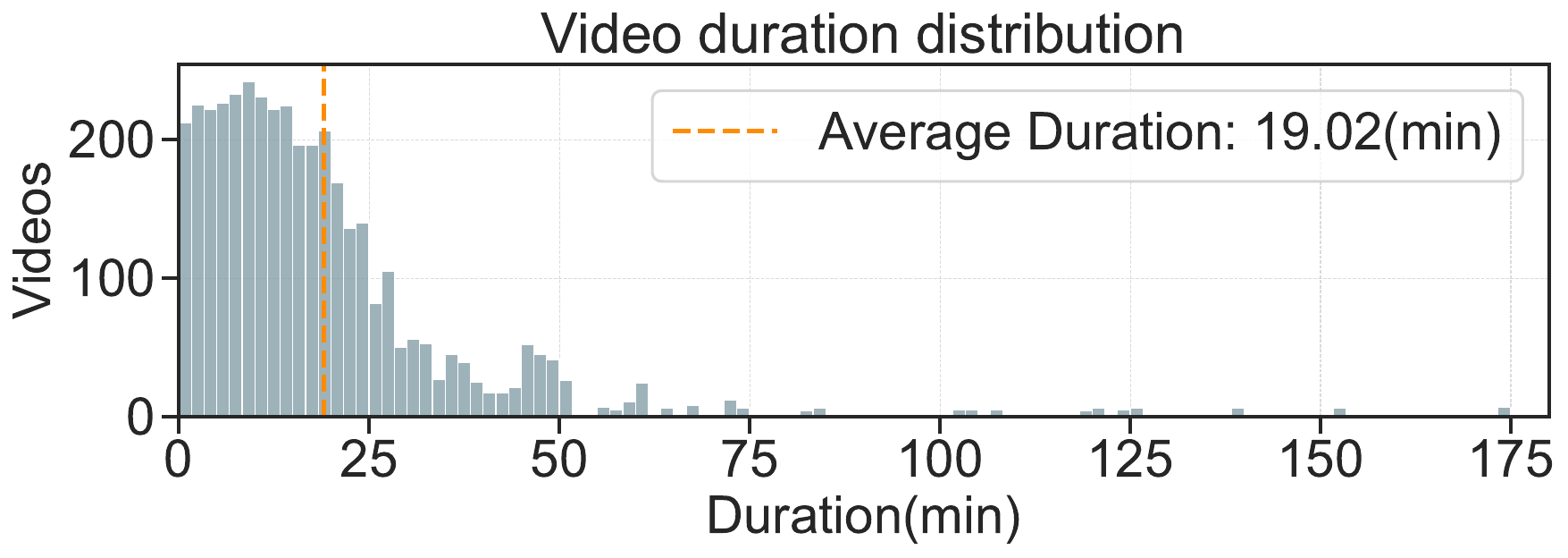}
  \caption{Distribution of video durations in SLVMEval. We divide the video duration into 100-second intervals and plot the number of videos included in each interval.}
  \label{fig:dataset_duration_distribution}
\end{figure}

Figure~\ref{fig:dataset_duration_distribution} shows the distribution of video durations.
We partition the videos into 100-second bins and plot the number of videos in each bin.
Consistent with the description in the main paper, the dataset covers a wide range of durations, from several minutes to very long videos of up to 10{,}486 seconds (approximately 2 hours and 54 minutes), with an average duration of 1{,}141 seconds (about 19 minutes).
It covers a wide range of video durations, from a few minutes to several hours, allowing us to evaluate the long-form videos targeted by SLVMEval.

\section{Details of degrading operations}
This section provides details on some of the aspects described in Section~\ref{subsec:degradation_operations}.

\subsection{Information Extraction from Captions}
\label{appendix:subsec:extract_info_from_caption}
As a preprocessing step, we extracted the information necessary for the degrading operations from the caption annotations of each video clip included in the source dataset (Vript~\cite{NEURIPS2024_6903a5aa}).
Specifically, we used Qwen3-8B (thinking mode)~\cite{qwen3technicalreport} to extract content mentioning object colors, background elements, and the dynamics degree.
We perform inference with vllm~\cite{kwon2023efficient} and use the structured outputs mode to enforce JSON format for the output.
The template for the prompt given to the model is as follows.

\begin{Verbatim}[breaklines=true,breaksymbolleft={}]
# Video Caption Analysis Prompt

## Task Description
Analyze the provided single scene video caption and extract information related to 7 evaluation criteria. Extract relevant information for each criterion and output in JSON format.

## Evaluation Criteria Definitions

- **subject**: Objects in the scene (e.g., person, animal, car, etc.)
  - Distinguish from scenes which refer to places or environments

- **background**: References to background elements

- **dynamic_degree**: Object movements within the caption (e.g., person running, jumping, car driving, bird flying)
  - Exclude camera movements (zoom, pan, etc.)
  - Record only object movements

- **human_action**: Human actions (e.g., running, jumping, walking, sitting)
  - Record only human actions
  - Use terms like "person", "man", "woman" for agents

- **color**: Color names only (e.g., red, white, blue, green)
  - For "red car", record "red" and "car" separately
  - Map colors to their corresponding objects

- **spatial_relationship**: Clear spatial relationships between objects (left, right, up, down, front, back, etc.)
  - Extract only when clear positions are stated like "upper right", "lower left"
  - Map positions to objects at those locations

- **scene**: Scene types or environments (e.g., river, ocean, city, mountain, etc.)
  - Refers to places or environments, distinct from individual objects

## Analysis Instructions

Analyze the caption in detail and execute the following:

1. **Information Extraction**: Extract specific information related to each evaluation criterion
2. **Empty Cases**: If no relevant information exists, output as empty list []

## Output Format

Output in the following JSON format:

```json
{
  "subject": ["object1", "object2"],
  "background": ["background_element1", "background_element2"],
  "dynamic_degree": [
    {
      "agent": "actor",
      "action": "action_description"
    }
  ],
  "human_action": [
    {
      "agent": "person",
      "action": "action_description"
    }
  ],
  "color": [
    {
      "color": "color_name",
      "object": "object_with_that_color"
    }
  ],
  "spatial_relationship": [
    {
      "location": "position (up, down, left, right, etc.)",
      "object": "object_at_that_position"
    }
  ],
  "scene": ["place1", "place2"]
}
```

## Analysis Hints

Keywords and expressions to look for in each evaluation criterion:

- **subject**: person, man, woman, child, animals (dog, cat, bird, etc.), car, bicycle, building, object names
- **background**: background, behind, backdrop, in the back
- **dynamic_degree**: 
  - Include: move, run, fly, flow, rotate, travel (object actions)
  - Exclude: zoom in, zoom out, pan, tilt (camera actions)
- **human_action**: run, walk, sit, stand, jump, speak, eat, read, write, point, touch
- **color**: 
  - Color names: red, blue, green, yellow, white, black, pink, purple, orange
  - Also record corresponding objects
- **spatial_relationship**: 
  - Clear positions only: left, right, up, down, front, back, upper left, upper right, lower left, lower right, center, below, above
  - Do not extract: next to, between, near, far (ambiguous expressions)
- **scene**: 
  - Places/environments: park, street, ocean, mountain, river, forest, indoor, outdoor, city, nature, tunnel, road, room
  - Not objects: car, building, tree (these are classified as subject)

## Important Notes

- Extract only information explicitly stated in the caption
- Avoid speculation or assumptions, record only certain information
- If no information exists, leave as empty list []

## Caption to Analyze

Analyze the following caption:
```txt
<<<|SCENE_CAPTION|>>>
```
\end{Verbatim}

We insert the caption (prompt) of Vript into the $\mathrm{<<<|SCENE\_CAPTION|>>>}$ placeholder.
We set the inference hyperparameters according to the officially recommended parameters~\cite{qwen3technicalreport}: top-p=0.95, top-k=20, temperature=0.6, and min-p=0.
Additionally, we set a repetition-penalty of 1.2 to suppress repetition in the output.

\subsection{Degradation algorithms}
We performed degrading operations corresponding to each evaluation aspect using the following procedures.

\paragraph{Aesthetics.}
To synthesize $v_{p}^{-}$, we apply FFmpeg's \texttt{eq} filter to five clips from the original video with the \texttt{contrast} option set to $-0.8$.
In this process, the luminance (luma, $Y$) component is first inverted ($Y \to 1 - Y$), and then its dynamic range is compressed to 80\%.
The resulting video $v_{p}^{-}$ has a dull, flat appearance with reduced contrast.

\paragraph{Technical Quality.}
To synthesize $v_{p}^{-}$, we first resize all frames of the original video to a unified base resolution where the longer side is 512\,px while preserving the aspect ratio.
We then downscale only the clips targeted for degradation to a longer side of 256\,px using LANCZOS resampling, again preserving the aspect ratio.
Finally, we upscale these clips back to the base resolution (longer side 512\,px).
As a result, $v_{p}^{-}$ has reduced sharpness and a loss of fine details.

\paragraph{Appearance Style.}
To synthesize $v_{p}^{-}$, we randomly select one of the following five appearance styles for each video: cartoon, detail enhancement, oil painting, colored pencil sketch, and watercolor.
We then apply the selected style transformation to all frames in the clips targeted for degradation.
Each style transformation is implemented using OpenCV's~\cite{opencv_library} stylization filters.
Details of each transformation are provided below.

\begin{itemize}[leftmargin=*, align=left]
  \item[\textbf{Cartoon Style.}]
  For the cartoon style transformation, we first smooth color regions using \texttt{cv2.edgePreservingFilter} (\texttt{flags}=1, \texttt{sigma\_s}=40, \texttt{sigma\_r}=0.20). We then convert the image to grayscale with \texttt{cv2.cvtColor} and apply \texttt{cv2.medianBlur} (kernel size of 5). Next, we generate an edge mask by applying \texttt{cv2.adaptiveThreshold} (maximum value of 255, \texttt{ADAPTIVE\_THRESH\_MEAN\_C}, \texttt{THRESH\_BINARY}, \texttt{blockSize}=11, \texttt{C}=3), followed by another \texttt{cv2.medianBlur} (kernel size of 5), and finally combine it with the smoothed image using \texttt{cv2.bitwise\_and}.

  \item[\textbf{Detail Enhancement.}]
  For the detail enhancement style, we apply \texttt{cv2.detailEnhance} (\texttt{sigma\_s}=5, \texttt{sigma\_r}=0.08) to each frame.

  \item[\textbf{Oil Painting Style.}]
  For the oil painting style transformation, we apply \texttt{cv2.xphoto.oilPainting} (\texttt{size}=5, \texttt{dynRatio}=1) to each frame.

  \item[\textbf{Colored Pencil Sketch Style.}]
  For the colored pencil sketch style transformation, we apply \texttt{cv2.pencilSketch} (\texttt{sigma\_s}=40, \texttt{sigma\_r}=0.05, \texttt{shade\_factor}=0.015).

  \item[\textbf{Watercolor Style.}]
  For the watercolor style transformation, we apply \texttt{cv2.stylization} (\texttt{sigma\_s}=40, \texttt{sigma\_r}=0.25) to each frame.
\end{itemize}

\begin{table*}[t]
\caption{
  Accuracy (\%) of CLIPScore under different prompt-handling strategies.
  Numbers are accuracy\,\% $\pm$ 95\% CI (percentage points).
  The chance level is 50\%.
}
  \centering
  \small
  \setlength{\tabcolsep}{1.0pt}
  \renewcommand{\arraystretch}{0.9}
  \begin{tabular*}{\textwidth}{@{\extracolsep{\fill}}l *{10}{C{0.08\textwidth}}}
    \toprule
    & \multicolumn{4}{c}{{\categoryfont Video Quality}}
    & \multicolumn{6}{c}{{\categoryfont Video-Text Consistency}} \\
    \cmidrule(lr){2-5}\cmidrule(lr){6-11}
    \noalign{\vspace{-1.5pt}}
    &
    \thead{Aesthetics} &
    \thead{Technical\\Quality} &
    \thead{Appearance\\Style} &
    \thead{Background\\Consistency} &
    \thead{Object\\Integrity} &
    \thead{Color} &
    \thead{Dynamics\\Degree} &
    \thead{Comprehen-\\siveness} &
    \thead{Spatial\\Relationship} &
    \thead{Temporal\\Flow} \\[-1.5pt]
    \midrule

    \lbl{IgnMax}
      & \accpm{48.9}{5.8}
      & \accpm{76.2}{7.3}
      & \accpm{62.8}{5.4}
      & \accpm{70.9}{4.7}
      & \accpm{71.0}{8.9}
      & \accpm{66.2}{6.5}
      & \accpm{55.3}{5.3}
      & \accpm{51.1}{6.4}
      & \accpm{60.2}{6.2}
      & \accpm{56.1}{5.8} \\

    \lbl{EachTrunc}
      & \accpm{54.3}{5.8}
      & \accpm{72.3}{7.7}
      & \accpm{66.0}{5.3}
      & \accpm{76.6}{4.4}
      & \accpm{75.0}{8.5}
      & \accpm{68.1}{6.4}
      & \accpm{55.0}{5.3}
      & \accpm{60.4}{6.3}
      & \accpm{56.4}{6.3}
      & \accpm{54.4}{5.8} \\

    \bottomrule
  \end{tabular*}
  \label{tab:clipscore_prompt_handling}
\end{table*}

\paragraph{Background Consistency.}
To synthesize $v_{p}^{-}$, we apply the following three-step procedure:
1. Randomly sample five clips from the original video,
2. Remove the background from all frames in each sampled clip,
3. Add a new background image to the background-removed frames.

In step 1, the sampling process is restricted to clips whose captions extracted by Qwen3 contain background-related information.
For step 2, we remove the background using the Python library rembg~\cite{rembg}, which internally uses $U^2$-Net~\cite{QIN2020107404}.
In step 3, we randomly sample new background images from the nature-dataset~\cite{nature_dataset_hf} and paste them onto the background-removed frames.
As a result, $v_{p}^{-}$ becomes a video whose background is inconsistent across adjacent frames.

\paragraph{Object Integrity.}
As described in Section~\ref{subsec:degradation_operations}, we degrade object integrity in two steps: (i) detecting the positions of objects mentioned in the caption and (ii) removing these objects by inpainting.
We randomly select target objects for deletion based on the information extracted from the caption (see Appendix~\ref{appendix:subsec:extract_info_from_caption}).
To obtain bounding boxes (bbox) of the target objects in each frame, we use IDEA-Research/grounding-dino-base~\cite{liu2023grounding}. 
We first resize each frame so that its longer side is 512\,px while preserving the aspect ratio, and then estimate the bounding boxes.
During detection, we set the thresholds to box-threshold=0.4 and text-threshold=0.3.

For object removal via inpainting, we use stable-diffusion-v1-5/stable-diffusion-inpainting~\cite{Rombach_2022_CVPR}. 
Similarly, we resize each frame so that its longer side is 512\,px while preserving the aspect ratio, and set denoising-steps to 50 and guidance-scale to 7.5.

\paragraph{Color.}
To synthesize $v_{p}^{-}$, we use Qwen-Image-Edit-2509~\cite{wu2025qwenimagetechnicalreport}, an image editing model to locally change only the colors of objects.

In this degradation procedure, we first randomly sample clips whose captions contain object color attribute information.
We extract this attribute information using the method described in Appendix~\ref{appendix:subsec:extract_info_from_caption}.
For every frame in a sampled clip, we use Qwen-Image-Edit to change the color of a target object from its original color to a new one (e.g., ``a red car'' $\to$ ``a blue car'').
We choose the new color by randomly sampling one color that is different from the original from a predefined set of colors (black, white, red, green, yellow, blue, brown, purple, pink, orange, gray).
We provide Qwen-Image-Edit with the prompt
\texttt{"Change the color of the \{object\} to \{color\}."}
to instruct it to change only the color of the specified object.

\begin{table*}[t]
    \caption{
         Correspondence between aspect, dimension, and description. The dimension and description are included in the prompt.
    }
    \centering
    \begin{tabular}{llp{0.5\linewidth}}
        \toprule
        aspect & dimension & description \\
        \midrule
        Aesthetics & aesthetics &
        This viewpoint indicates whether each frame of the video exhibits overall visual appeal, including composition, harmony, and photographic quality. \\
        \midrule
        Technical Quality & technical\_quality &
        This viewpoint indicates whether each frame is free from technical artifacts (e.g., blur, noise, compression artifacts, over- or under-exposure). \\
        \midrule
        Appearance Style & appearance\_style &
        This viewpoint indicates whether the required appearance style in the text prompt (e.g., oil painting style, black-and-white style, watercolor style, cyberpunk style) is appropriately represented. \\
        \midrule
        Background Consistency & background\_consistency &
        This viewpoint indicates whether the background remains visually stable and consistent across frames. \\
        \midrule
        Object Integrity & object\_removal &
        This viewpoint indicates whether a specific object (e.g., dog, car) mentioned in the text prompt is correctly generated. \\
        \midrule
        Color & color &
        This viewpoint indicates whether the colors specified in the text prompt (for instance, the color of an object) are correctly reproduced. \\
        \midrule
        Dynamics Degree & dynamics\_degree &
        This viewpoint indicates whether the generated video contains an appropriate amount of motion (neither too static nor overly erratic). \\
        \midrule
        Comprehensiveness & scene &
        This viewpoint indicates whether the overall scene (for example, “ocean” when prompted with “ocean”) is generated correctly according to the text prompt. \\
        \midrule
        Spatial Relationship & spatial\_relationship &
        This viewpoint indicates whether generated objects follow the text prompt's spatial instructions across four relation types: Horizontal, Vertical, Proximity/Adjacency, and Depth/Occlusion. \\
        \midrule
        Temporal Flow & move\_scene &
        This viewpoint assesses whether the generated video preserves a coherent, chronological flow of scenes, avoiding unintended shuffling, repetition, or time-jumps-in line with the order implied by the text prompt. \\
        \bottomrule
    \end{tabular}
    \label{tab:aspect-to-dimension-to-description}
\end{table*}

\section{Effectiveness of our frame sampling strategy}
\label{appendix:sec:sampling_effectiveness}

Many current evaluation systems impose strict input limits (e.g., maximum frames or context length), making it impractical to feed all frames of long videos.
Since SLVMEval explicitly targets videos that range from minutes to hours, a sampling strategy is required to make evaluation feasible while retaining coverage over the full temporal span.

\subsection{Our sampling design}
We adopt a clip-based sampling strategy: we detect clip boundaries using FFmpeg and extract the center frame from each detected clip (described in \S5.1 and \S5.2).
If the resulting frame sequence still exceeds the evaluation system's maximum frame budget, we randomly subsample frames to satisfy the limit.
This strategy is widely used in long-video understanding to improve temporal coverage under limited budgets~\cite{qian2024streaming,zeng2025timesuite}.

The key rationale is that uniform per-clip sampling spreads the budget across the entire video, reducing the risk that the evaluation systems sees only early segments.
This is particularly important for SLVMEval, because degradations can appear at arbitrary temporal positions.

\subsection{Sequence length limits}
We evaluate CLIPScore under different prompt-length handling policies to examine whether our conclusions are sensitive to our sampling design.
In addition to the default setting(\S5.2), we consider two alternatives:
(i) \lbl{IgnMax}; which ignores the model’s maximum sequence length \footnote{We cap the input at 31{,}000 tokens due to GPU memory constraints}
and (ii) \lbl{EachTrunc}; which evenly truncates each clip caption so that the final concatenated prompt fits within the model’s maximum length.
Per-aspect results are reported in Table~\ref{tab:clipscore_prompt_handling}. 

As a result, the Average CLIPScore accuracy (\%) across all aspects was \lbl{Default} (\S 5.2): 60.7\%, \lbl{IgnMax}: 60.5\%, and \lbl{EachTrunc}: 62.8\%, indicating no substantial differences.

\section{Details of baseline systems}
\label{appendix:sec:baseline_systems}

\subsection{Pairwise VLM-as-a-judge}
We used three VLMs: GPT-5 (gpt-5-2025-08-07), GPT-5-mini (gpt-5-mini-2025-08-07), and Qwen3 (Qwen/Qwen3-VL-235B-A22B-Thinking-FP8).
For all models, we performed inference using the Structured Outputs mode to enforce JSON-formatted outputs.
We set the inference hyperparameters for Qwen3-VL to top-p = 0.8, top-k = 20, temperature = 0.7, and min-p = 0, and for GPT-5 we used temperature = 1.0.\footnote{We avoided greedy decoding for Qwen3 inference because it is deprecated to prevent repetition. Additionally, for GPT-5 we used a temperature of 1.0 because it is the only supported and fixed value.}

\begin{algorithm}[t]
  \caption{Implementation of Video-based evaluation}
  \label{alg:video_only_eval_simple}
  \begin{algorithmic}[1]
    \Require Prompt $p$, video $u_{p}$, video $v_{p}$, aspect $a$, maximum number of frames $F_{\mathrm{max}}$
    \Ensure $u_{p}$ or $v_{p}$
    \State $u_{p}^{s} \gets \textsc{SampleFrames}(u_{p})$
    \State $v_{p}^{s} \gets \textsc{SampleFrames}(v_{p})$
    \For{$s \in \{u_{p}^{s
    }, v_{p}^{s}\}$}
      \If{$|s| > F_{\mathrm{max}}$}
        \State $s \gets \textsc{RandomSample}(s, F_{\mathrm{max}})$ 
      \EndIf
    \EndFor
    \State $d \gets \textsc{VLM}(a, p, u_{p}^{s}, v_{p}^{s})$ \Comment{Return ``first'' or ``second''}
    \If{$d = \text{``first''}$}
      \State \Return $u_{p}^{s}$
    \ElsIf{$d = \text{``second''}$}
      \State \Return $v_{p}^{s}$
    \EndIf
  \end{algorithmic}
\end{algorithm}

\begin{algorithm}[t]
  \caption{Implementation of Text-based evaluation}
  \label{alg:text_based_eval}
  \begin{algorithmic}[1]
    \Require Prompt $p$, video $u_{p}$, video $v_{p}$, aspect $a$, maximum number of frames $F_{\mathrm{max}}$
    \Ensure $u_{p}$ or $v_{p}$
    \State $u_{p}^{s} \gets \textsc{SampleFrames}(u_{p})$
    \State $v_{p}^{s} \gets \textsc{SampleFrames}(v_{p})$
    \For{$s \in \{u_{p}^{s}, v_{p}^{s}\}$}
      \If{$|s| > F_{\mathrm{max}}$}
        \State $s \gets \textsc{RandomSample}(s, F_{\mathrm{max}})$ 
      \EndIf
    \EndFor
    
    \State $caption_{u_p} \gets \mathrm{VLM}_{\mathrm{cap}}(a, u_p^{s})$
    \State $caption_{v_p} \gets \mathrm{VLM}_{\mathrm{cap}}(a, v_p^{s})$
    
    \State $d \gets \textsc{LM}(a, p, caption_{u_{p}}, caption_{v_{p}})$
    \Comment{Return ``first'' or ``second''}
    
    \If{$d = \text{``first''}$}
      \State \Return $u_{p}$
    \ElsIf{$d = \text{``second''}$}
      \State \Return $v_{p}$
    \EndIf
  \end{algorithmic}
\end{algorithm}

\begin{table*}[t]
\caption{
  Spearman rank correlation coefficients $\rho_{\mathrm{S}}$ between video duration and accuracy for each evaluation system (rows) and evaluation aspect (columns).
  For each aspect and system, we sort the test samples by video duration, divide them into 50 bins, compute the accuracy within each bin, and then compute Spearman's $\rho_{\mathrm{S}}$ between the bin index (corresponding to video duration) and the bin-wise accuracy.
  Negative values indicate that accuracy tends to decrease as video duration increases.
}
  \centering
  \small
  \setlength{\tabcolsep}{1.0pt}
  \renewcommand{\arraystretch}{0.9}
  \begin{tabular*}{\textwidth}{@{\extracolsep{\fill}}l *{10}{C{0.08\textwidth}}}
    \toprule
    & \multicolumn{4}{c}{{\categoryfont Video Quality}}
    & \multicolumn{6}{c}{{\categoryfont Video-Text Consistency}} \\
    \cmidrule(lr){2-5}\cmidrule(lr){6-11}
    \noalign{\vspace{-1.5pt}}
    &
    \thead{Aesthetics} &
    \thead{Technical\\Quality} &
    \thead{Appearance\\Style} &
    \thead{Background\\Consistency} &
    \thead{Object\\Integrity} &
    \thead{Color} &
    \thead{Dynamics\\Degree} &
    \thead{Comprehen-\\siveness} &
    \thead{Spatial\\Relationship} &
    \thead{Temporal\\Flow} \\[-1.5pt]
    \midrule

    Video-based \\
    \quad GPT-5
      & -0.6806 & -0.4930 & -0.1150 &  0.0487 &  0.0541 & -0.1198 &  0.1446 & -0.2200 & -0.1216 & -0.4927 \\
    \quad GPT-5-mini
      & -0.5422 & -0.3558 & -0.2473 & -0.2874 & -0.2409 & -0.3055 & -0.0402 & -0.2239 & -0.1750 & -0.3623 \\
    \quad Qwen3
      & -0.1368 & -0.0572 & -0.0636 & -0.2014 &  0.3848 & -0.0683 &  0.1418 &  0.0200 & -0.2934 &  0.1239 \\

    Text-based \\
    \quad GPT-5
      & -0.4650 & -0.1872 & -0.3206 & -0.5072 & -0.2892 & -0.3270 &  0.1386 & -0.3076 & -0.0776 & -0.4547 \\
    \quad GPT-5-mini
      & -0.4985 & -0.2639 & -0.3791 & -0.6065 & -0.2899 & -0.3620 &  0.1504 & -0.2847 & -0.0541 & -0.4107 \\
    \quad Qwen3
      & -0.0713 & -0.2954 & -0.4476 & -0.6920 & -0.1338 & -0.5632 &  0.0020 & -0.0061 & -0.3756 & -0.4559 \\

    CLIPScore
      & -0.2555 & -0.0218 & -0.2681 & -0.4483 & -0.2538 & -0.7167 & -0.0716 & -0.2165 & -0.1322 & -0.1484 \\
    VideoScore
      &  0.1677 & -0.0094 & -0.4636 & -0.2936 & -0.3749 & -0.4406 & -0.2625 & -0.4824 &  0.3212 & -0.1139 \\
    Human
      & -0.3847 & -0.3902 &  0.2202 &  0.2360 & -0.1905 &  0.0158 & -0.4254 & -0.3855 & -0.4621 &  0.0845 \\
    \bottomrule
  \end{tabular*}
  \label{tab:duration_corr}
\end{table*}

\begin{table*}[t]
\caption{
  Two-sided $p$-values for the Spearman rank correlations in Table~\ref{tab:duration_corr}.
  Each entry is rounded to four decimal places.
  Bold values indicate statistically significant correlations with $p < 0.05$.
}
  \centering
  \small
  \setlength{\tabcolsep}{1.0pt}
  \renewcommand{\arraystretch}{0.9}
  \begin{tabular*}{\textwidth}{@{\extracolsep{\fill}}l *{10}{C{0.08\textwidth}}}
    \toprule
    & \multicolumn{4}{c}{{\categoryfont Video Quality}}
    & \multicolumn{6}{c}{{\categoryfont Video-Text Consistency}} \\
    \cmidrule(lr){2-5}\cmidrule(lr){6-11}
    \noalign{\vspace{-1.5pt}}
    &
    \thead{Aesthetics} &
    \thead{Technical\\Quality} &
    \thead{Appearance\\Style} &
    \thead{Background\\Consistency} &
    \thead{Object\\Integrity} &
    \thead{Color} &
    \thead{Dynamics\\Degree} &
    \thead{Comprehen-\\siveness} &
    \thead{Spatial\\Relationship} &
    \thead{Temporal\\Flow} \\[-1.5pt]
    \midrule

    Video-based \\
    \quad GPT-5
      & $\mathbf{0.0000}$ & $\mathbf{0.0003}$ & $0.4263$ & $0.7371$ & $0.7090$ & $0.4073$ & $0.3164$ & $0.1248$ & $0.4002$ & $\mathbf{0.0003}$ \\
    \quad GPT-5-mini
      & $\mathbf{0.0000}$ & $\mathbf{0.0112}$ & $0.0834$ & $\mathbf{0.0430}$ & $0.0919$ & $\mathbf{0.0310}$ & $0.7818$ & $0.1181$ & $0.2241$ & $\mathbf{0.0097}$ \\
    \quad Qwen3
      & $0.3434$ & $0.6934$ & $0.6606$ & $0.1607$ & $\mathbf{0.0058}$ & $0.6373$ & $0.3259$ & $0.8902$ & $\mathbf{0.0387}$ & $0.3915$ \\

    Text-based \\
    \quad GPT-5
      & $\mathbf{0.0007}$ & $0.1929$ & $\mathbf{0.0232}$ & $\mathbf{0.0002}$ & $\mathbf{0.0416}$ & $\mathbf{0.0205}$ & $0.3369$ & $\mathbf{0.0298}$ & $0.5922$ & $\mathbf{0.0009}$ \\
    \quad GPT-5-mini
      & $\mathbf{0.0002}$ & $0.0640$ & $\mathbf{0.0066}$ & $\mathbf{0.0000}$ & $\mathbf{0.0411}$ & $\mathbf{0.0098}$ & $0.2970$ & $\mathbf{0.0451}$ & $0.7090$ & $\mathbf{0.0030}$ \\
    \quad Qwen3
      & $0.6226$ & $\mathbf{0.0373}$ & $\mathbf{0.0011}$ & $\mathbf{0.0000}$ & $0.3542$ & $\mathbf{0.0000}$ & $0.9889$ & $0.9664$ & $\mathbf{0.0072}$ & $\mathbf{0.0009}$ \\

    CLIPScore
      & $0.0733$ & $0.8806$ & $0.0598$ & $\mathbf{0.0011}$ & $0.0753$ & $\mathbf{0.0000}$ & $0.6213$ & $0.1311$ & $0.3601$ & $0.3036$ \\
    VideoScore
      & $0.2443$ & $0.9484$ & $\mathbf{0.0007}$ & $\mathbf{0.0385}$ & $\mathbf{0.0073}$ & $\mathbf{0.0014}$ & $0.0656$ & $\mathbf{0.0004}$ & $\mathbf{0.0229}$ & $0.4311$ \\
    Human
      & $\mathbf{0.0058}$ & $\mathbf{0.0051}$ & $0.1244$ & $0.0990$ & $0.1851$ & $0.9132$ & $\mathbf{0.0021}$ & $\mathbf{0.0057}$ & $\mathbf{0.0007}$ & $0.5595$ \\
    \bottomrule
  \end{tabular*}
  \label{tab:duration_corr_p}
\end{table*}

\section{Correlation between accuracy and video duration}
\label{sec:correlation_duration}

In the main paper (Figure~\ref{fig:degrade_pos_vs_acc}), we analyzed how the accuracy of each evaluation system changes as a function of video duration.
Here we provide a more detailed analysis based on Spearman rank correlation coefficients $\rho_{\mathrm{S}}$ between video duration and accuracy for each evaluation aspect and each evaluation system.

For every aspect--system pair, we follow the procedure described in the caption of Figure~\ref{fig:degrade_pos_vs_acc}.
Table~\ref{tab:duration_corr} reports the resulting $\rho_{\mathrm{S}}$ values, and Table~\ref{tab:duration_corr_p} reports the corresponding two-sided $p$-values.

Overall, most automatic evaluation systems exhibit negative correlations on many aspects, indicating that their accuracy tends to decrease as video duration increases.
This tendency is especially pronounced for aspects such as \textit{Background Consistency}, \textit{Color}, and \textit{Temporal Flow}, where several systems show relatively large negative $\rho_{\mathrm{S}}$ with statistically significant $p$-values ($p < 0.05$).
In contrast, for aspects such as \textit{Dynamics Degree}, many systems already perform near chance level for both short and long videos.
As a result, the correlations are small and often not statistically significant.

We also report the correlation values for human evaluators.
Humans show only weakly negative correlations, and their $p$-values are generally larger than those of the automatic systems, reflecting that human performance remains high and relatively stable across different video durations.
This further highlights the gap between human robustness and the current limitations of automatic evaluation systems for T2LV generation.

\paragraph{Video-based evaluation.}
The prompt template for the video-based evaluation is as follows.
The \texttt{input\_text} placeholder holds the input prompt for the T2V model.
See Section~\ref{sec:aspect-to-dimension-to-description} for the text used for the \texttt{dimension} and \texttt{description} placeholders.

\begin{Verbatim}[breaklines=true,breaksymbolleft={}]
[System prompt]
You are a meticulous AI evaluator for video. 
You will be provided with the evaluation viewpoint and its description, the text used for generation, and frames of the videos generated by two different videos.
Based on the given viewpoint, compare the quality of the two videos in relation to the input text.
Finally, output which video, 'first video' or 'second video', has the higher quality in JSON format.

[User prompt]
Given the provided information below, please evaluate which video, the first video or the second video, generated a higher-quality video.

Viewpoints and descriptions: '''
- {dimension}: {description}
'''

Text input used for generation: '''
{input_text}
'''
\end{Verbatim}

Algorithm~\ref{alg:video_only_eval_simple} presents the pseudocode for the entire procedure.

\begin{table*}[t]
  \caption{Computing resources used in our experiments.}
  \centering
  \begin{tabular}{lll}
    \toprule
     & \textbf{CPU resource} & \textbf{GPU resource} \\
    \midrule
    CPU Model & \makecell[l]{2 x Intel Xeon Platinum 8490H\\(Sapphire Rapids)} & \makecell[l]{2 x Intel Xeon Platinum 8490H\\(Sapphire Rapids)} \\
    GPU Model & None & 4 x NVIDIA H100 (HBM2e) \\
    System Memory (per node) & 512 GiB (DDR5) & 1 TiB (DDR5) \\
    GPU Memory (per node) & N/A & 94 GiB per GPU \\
    Operating System & Rocky Linux & Rocky Linux \\
    \bottomrule
  \end{tabular}
  \label{tab:computing_resource}
\end{table*}

\paragraph{Text-based evaluation.}
The prompt template for the text-based evaluation is as follows.
The \texttt{input\_text} placeholder contains the input prompt for the T2V model, while the \texttt{first\_model\_text} and \texttt{second\_model\_text} placeholders hold the generated captions for the respective videos. 
See Section~\ref{sec:aspect-to-dimension-to-description} for the text used for the \texttt{dimension} and \texttt{description} placeholders.

\begin{Verbatim}[breaklines=true,breaksymbolleft={}]
[System prompt]
You are a meticulous AI evaluator for video. 
You will be provided with the evaluation viewpoint and its description, the text used for generation, and transcriptions of the videos generated by two different videos.
Based on the given viewpoint, compare the quality of the two videos in relation to the input text.
Finally, output which video, 'first video' or 'second video', has the higher quality in JSON format.

[User prompt]
Given the provided information below, please evaluate which video, the first video or the second video, generated a higher-quality video.

Viewpoints and descriptions: '''
- {dimension}: {description}
'''

Text input used for generation: '''
{input_text}
'''

Transcription of the video generated by the first video: '''
{first_model_text}
'''

Transcription of the video generated by the second video: '''
{second_model_text}
'''
\end{Verbatim}

Algorithm~\ref{alg:text_based_eval} presents the pseudocode for the entire procedure.

\paragraph{Captioning.}
The prompt we use to generate caption in the text-based and hybrid evaluations is as follows.

\begin{Verbatim}[breaklines=true,breaksymbolleft={}]
[System prompt]
You are an assistant that converts video content into a thorough textual description.
You will be given one or more frames from a video (or a description of them).
Your goal is to accurately describe all visible details and actions, without adding information that cannot be inferred from the video.
For each provided evaluation viewpoint, please represent the video content as text and output the result in JSON format.

Important notes: '''
- Only include details that can be directly observed from the given frames.
- Do not speculate or assume any information not clearly visible.
- Describe objects, people, scenes, backgrounds, movements, aesthetics, video quality and any relevant visual details.
- Be concise and factual. Avoid subjective or interpretive language.
- If multiple frames are provided, indicate changes or continuity across them.
- Output your description in a structured JSON format.
'''

Viewpoints and descriptions: '''
Viewpoint: {aspect}
Description: {description}
'''

[User prompt]
Please generate descriptions based on the evaluation viewpoints for the given videos.
\end{Verbatim}

\subsection{Prompt component mapping}
\label{sec:aspect-to-dimension-to-description}
Table~\ref{tab:aspect-to-dimension-to-description} shows the correspondence between the evaluation aspect and the \texttt{dimension} and \texttt{description} text provided to the evaluation system.

\subsection{VideoScore}
As described in Section~\ref{subsec:videoscore}, we use VideoScore-v1.1~\cite{he-etal-2024-videoscore} as one of our baseline evaluation systems.
Because the aspects we consider differ from those in VideoScore, we construct a mapping between our aspects and those of VideoScore.
This mapping is summarized in Table~\ref{tab:video_score_aspect_mapping}.

\begin{table}[t]
  \caption{Mapping between our defined aspects and those defined in VideoScore.}
  \centering
  \begin{tabular}{l l}
    \hline
    Our aspect             & VideoScore aspect        \\
    \hline
    Aesthetics             & visual quality           \\
    Technical Quality      & visual quality           \\
    Appearance Style       & visual quality           \\
    Background Consistency & factual consistency      \\
    Object Integrity       & text-to-video alignment  \\
    Color                  & visual quality           \\
    Dynamics Degree        & dynamic degree           \\
    Comprehensiveness      & text-to-video alignment  \\
    Spatial Relationship   & text-to-video alignment  \\
    Temporal Flow          & text-to-video alignment  \\
    \hline
  \end{tabular}
  \label{tab:video_score_aspect_mapping}
\end{table}

\section{Computational Resources}
We used "Genkai," a computational resource provided by Kyushu University, for our experiments.
Table~\ref{tab:computing_resource} details the computational resources we use.

\section{Ethics statement}
We outsourced data annotations to crowd workers on Yahoo! Crowdsourcing.
Before participation, we explained the purpose of the task to all workers and obtained their informed consent.
We did not collect any personally identifiable information, and we paid workers at rates above the legal minimum wage in the country where the platform operates.

\section{Limitations}
The proposed SLVMEval benchmark benchmark utilizes a synthetic dataset; thus, its distribution differs from the outputs expected from future T2LV models and may not cover all failures that those models could produce.
Therefore, additional benchmarks are required to evaluate models under more complex and realistic settings.
Such benchmarks will need to be developed in parallel with advancements in T2LV generation models because their design will depend on analyzing model failure patterns.
Thus, we anticipate further progress in this research area.

In addition, the settings of the degrading operations, e.g., the magnitude of the contrast changes and the fraction of clips degraded, function as hyperparameters, and
varying these hyperparameters is expected to change the difficulty of the benchmark.
Currently, the benchmark is easy for humans to judge, which suits the evaluation goal considered in the current study; however, creating variants with different difficulty levels would enable broader analysis.

\end{document}